\documentclass[preprint,12pt]{elsarticle}

\usepackage{amssymb}

\usepackage{multirow}
\usepackage{comment}
\usepackage{amsmath,amssymb,amsfonts}
\usepackage{algorithmic}
\usepackage{graphicx}
\usepackage{textcomp}
\usepackage{xcolor}
\usepackage{hyperref}

\journal{Neurocomputing}

\begin{document}

\begin{frontmatter}



\title{Depth-Wise Convolutions in Vision Transformers for Efficient Training on Small Datasets}


\author[label1]{Tianxiao Zhang}
\author[label2]{Wenju Xu}
\author[label1]{Bo Luo}
\author[label3]{Guanghui Wang\corref{cor1}}
\cortext[cor1]{Corresponding author (wangcs@torontomu.ca)}

\affiliation[label1]{organization={Department of Electrical Engineering and Computer Science, University of Kansas},
            city={Lawrence},
            postcode={66045}, 
            state={KS},
            country={USA}}
\affiliation[label2]{organization={Amazon},
            city={Palo Alto},
            postcode={94301}, 
            state={CA},
            country={USA}}
\affiliation[label3]{organization={Department of Computer Science, Toronto Metropolitan University},
            city={Toronto},
            postcode={M5B 2K3}, 
            state={ON},
            country={Canada}}

\begin{abstract}
The Vision Transformer (ViT) leverages the Transformer's encoder to capture global information by dividing images into patches and achieves superior performance across various computer vision tasks. However, the self-attention mechanism of ViT captures the global context from the outset, overlooking the inherent relationships between neighboring pixels in images or videos. Transformers mainly focus on global information while ignoring the fine-grained local details. Consequently, ViT lacks inductive bias during image or video dataset training. In contrast, convolutional neural networks (CNNs), with their reliance on local filters, possess an inherent inductive bias, making them more efficient and quicker to converge than ViT with less data. 
In this paper, we present a lightweight Depth-Wise Convolution module as a shortcut in ViT models, bypassing entire Transformer blocks to ensure the models capture both local and global information with minimal overhead. Additionally, we introduce two architecture variants, allowing the Depth-Wise Convolution modules to be applied to multiple Transformer blocks for parameter savings, and incorporating independent parallel Depth-Wise Convolution modules with different kernels to enhance the acquisition of local information. The proposed approach significantly boosts the performance of ViT models on image classification, object detection, and instance segmentation by a large margin, especially on small datasets, as evaluated on CIFAR-10, CIFAR-100, Tiny-ImageNet and ImageNet for image classification, and COCO for object detection and instance segmentation. The source code can be accessed at \url{https://github.com/ZTX-100/Efficient_ViT_with_DW}.

\end{abstract}



\begin{keyword}
Vision Transformers, Depth-Wise Convolutions, Small Dataset

\end{keyword}

\end{frontmatter}


\section{Introduction}
Transformer models have demonstrated exceptional performance in Natural Language Processing (NLP) tasks by capturing long-range relationships through attention mechanisms \cite{vaswani2017attention}. However, the direct application of Transformer models to vision tasks is less intuitive, as images are inherently interconnected, and pixels exhibit close relationships. Vision Transformer (ViT) \cite{dosovitskiy2020image} addresses this challenge by dividing the image into fixed-size patches, linearly embedding each patch as a token. To capture 2D relationships among image tokens, positional embedding is introduced, compensating for the loss of 2D coordinate relationships in embedded image patches. ViT includes a learnable class token to interact with image patch tokens for image classification.

Despite its success, ViT often requires substantial data and longer training times due to the attention mechanism's computational demands. The attention mechanism calculates the dot product of embeddings for each token pair, necessitating more time to learn the inductive bias that neighboring pixels share stronger relationships. Global attention in ViTs treats all tokens equally, neglecting the fact that neighboring image patches have higher relationships. In contrast, Convolutional Neural Networks (CNNs) naturally possess inductive bias due to local filters. However, CNNs may have a lower upper bound than ViTs because of their limited global view. In essence, ViTs outperform CNNs when datasets are large enough, and training times are sufficiently long, showcasing their superior performance under such conditions.

Image contents are inherently cohesive as a whole, and forcefully splitting them into patches can hinder the recognition process. Moreover, treating all patches equally in models like Vision Transformers (ViTs) sacrifices the inductive bias present in the images, requiring a more extensive training effort to converge. While some approaches involve overlapping patches, this introduces additional computational costs without fundamentally addressing the issue. In contrast, CNN models, by their nature, excel at filtering local pixels in a contiguous manner, which is crucial for image recognition, particularly when dealing with relatively small objects. However, the lack of global views may restrict the performance of convolutional models, especially in scenarios with abundant training data. The key question becomes: How can we efficiently integrate these two approaches, leveraging convolutions to support Transformer models, ensuring rapid convergence, and achieving superior performance?

In this paper, we introduce a straightforward yet effective method to seamlessly integrate convolutional and Transformer blocks, enabling the simultaneous learning of global and local information efficiently. Our approach leverages Depth-Wise Convolutions \cite{howard2017mobilenets} to capture local information, while Transformer blocks are employed to capture global information. The Depth-Wise Convolutions serve as a shortcut, bypassing the entire Transformer block (attention+FFN). The final combination is achieved through summation, providing a unified representation of both Depth-Wise Convolutions and Transformer blocks. The Depth-Wise Convolutions are applied for each Transformer block, creating two paths after each block for the network to choose from. This design ensures a flexible and dynamic integration of the local and global features. Our method achieves a superior performance improvement with only a marginal increase in parameters and computations, particularly benefiting small datasets. Our approach enables small-size Transformer models to outperform some larger counterparts, showcasing their effectiveness and efficiency. 

In summary, our contributions are outlined below.

\begin{itemize}
    \item 
    We propose an efficient and effective approach to combining Depth-Wise Convolutions and Transformer blocks, allowing simultaneous capture of local and global information with minimal additional parameters and computational load. The proposed lightweight Depth-Wise module bypasses entire Transformer blocks to attain fine-grained details that might be missed otherwise. This module does not alter the internal structure of MHSA and FFN, making it a plug-and-play component that can be utilized by most Transformer models. Our approach demonstrates superior performance in image classification, object detection, and instance segmentation.
    \item  We developed two types of architectural variants. The first variant aims to reduce parameters and floating-point operations (FLOPs) by utilizing the Depth-Wise module to bypass multiple Transformer blocks. The second variant seeks to improve performance by incorporating multiple independent parallel Depth-Wise modules, each dedicated to enhancing local information.
    
    \item We demonstrate that certain modules are dispensable when our approach is implemented in the training of Transformer models on small datasets. Furthermore, by applying our approach without these modules, we can reduce both parameters and FLOPs, while significantly enhancing the performance.

\end{itemize}

\section{Related Work}
\subsection{Vision Transformers}
ViT \cite{dosovitskiy2020image} introduces the Transformer models into vision recognition by splitting the images into fixed-size patches and then tokenizing each patch into the token so that the image patches can be utilized in the attention module of Transformer models. Many variations and improvements have been proposed \cite{patel2022aggregating}\cite{chen2023accumulated}\cite{zhu2023biformer} and applied to various vision tasks such as point cloud completion \cite{wang2023pst} and crowd counting \cite{sajid2021audio}. DeiT \cite{touvron2021training} employs distillation tokens for attention learning from the teacher models to the student models. CaiT \cite{touvron2021going} introduces LayerScale to effectively train the ViT models with deeper layers so that the performance of deep ViT models could be further boosted. Hierarchical Vision Transformer architecture \cite{wang2021pyramid}\cite{heo2021rethinking}\cite{chen2021psvit}\cite{chen2022improving}\cite{chen2024superlora} are designed to better suit vision tasks by reducing the size of feature maps as the network progresses deeper, resembling the structure of CNN architectures.

To reduce the computational cost, some window-based Vision Transformer models have been proposed. Swin Transformer \cite{liu2021swin} restricts the self-attention of the tokens on small windows so that the inductive bias could be slightly introduced while significantly reducing the computational costs with the sacrifice of the global views. To mitigate the limitation of lacking global views, it also incorporates a shifted window mechanism, expanding the self-attention calculation to new shifted windows. Thus, the views of tokens are expanded. Other works, such as \cite{fang2022msg}\cite{chu2021twins}\cite{huang2021shuffle}, attempt to increase the receptive fields with cross-window interactions so that the information between the windows could be exchanged and the tokens could exchange the information with other windows.

\subsection{Vision Transformers and Convolutions}

CvT \cite{wu2021cvt} designs a hierarchical Transformer architecture with a convolutional token embedding and a convolutional Transformer block utilizing a convolution projection to project the feature maps into query, key, and value. BoTNet \cite{srinivas2021bottleneck} replaces the final three bottleneck blocks of the ResNet model with BoT blocks that contain MHSA layers so that the self-attention layer can aggregate the information attained by the convolutional layers. LocalViT \cite{li2021localvit} introduces the Depth-Wise Convolution into the Feed-Forward Networks in the Transformer block to add locality into the Transformer models. CMT \cite{guo2022cmt} proposes a hybrid Transformer model to take advantage of Transformers and CNNs for global views and local features, respectively. MobileFormer \cite{chen2022mobile} designs efficient networks to integrate MobileNet \cite{howard2017mobilenets} and Transformer blocks with a two-way bridge in between so that both local features and global interactions can be effectively communicated and fused. 

DHVT \cite{lu2022bridging} integrates the convolutions into MLP and patch embeddings to introduce the inductive bias into the Transformer model, and introduces a dynamic feature aggregation module in MLP and a "head token" in MHSA for diverse channel representation so that the gap between the Transformer models and CNN models could be eliminated. ViTAE \cite{xu2021vitae} and its extension model ViTAEv2 \cite{zhang2023vitaev2} utilize multiple dilated convolutions to downsample the feature maps and aid the MHSA module to attain the locality simultaneously. Mixformer \cite{chen2022mixformer} parallelizes window-based self-attention and Depth-Wise Convolution to extend the receptive fields and designs bi-directional interactions to exchange information of channel and spatial dimensions between them. DMFormer \cite{wei2023dmformer} proposes a Dynamic Multi-level Attention mechanism which is comprised of Depth-Wise Convolutions with multiple kernel sizes for various patterns and a gating mechanism for adaptability. ScopeViT \cite{nie2024scopevit} involves Depth-Wise Convolutions into Transformer architecture for scale-aware efficient training. DctViT \cite{su2024dctvit} proposes a hybrid structure with convolutions and Transformers for higher accuracy on multiple vision tasks. The hybrid structures are also applied to wetland classification \cite{jamali2023wetmapformer}, salient object detection \cite{wang2022hybrid}\cite{liu2023transcending}, referring image segmentation \cite{liu2023local}, etc.

The computational structures of some previous models focus on integrating convolutional networks into the Multi-Head Self-Attention (MHSA) or Feed-Forward Network (FFN), making convolutional networks essential components of Transformer architectures. Additionally, the convolutional components in some of these studies are not necessarily lightweight. In contrast, our approach aims to efficiently combine Transformer blocks with convolutions while minimizing computational overhead. It is designed as a versatile and straightforward module that can be easily integrated into various Vision Transformer models.

\section{Methodology}

\subsection{Vision Transformers}

ViT \cite{dosovitskiy2020image} introduces Transformers \cite{vaswani2017attention} into vision tasks by splitting the images into patches which are tokenized into tokens ($\mathbf{x}$). To preserve the positional relations of the image patches, learnable positional embeddings are added to each token to learn the 2D relations of the patches. The tokens and positional embeddings are illustrated in Eq.~\eqref{eq:1}. $\mathbf{x_c}$ demonstrates the class token and $\mathbf{x_p}$ indicates the positional embeddings.

\begin{equation}\label{eq:1}
    \mathbf{x^0} = (\mathbf{x_1}, \mathbf{x_2}, ..., \mathbf{x_l}; \mathbf{x_c}) + \mathbf{x_p}
\end{equation}

\begin{equation}\label{eq:2}
    \mathbf{x'^n} = \mathbf{x^n} + \text{MHSA}(\text{LayerNorm}(\mathbf{x^n}))
\end{equation}

\begin{equation}\label{eq:3}
    \mathbf{x^{n+1}} = \mathbf{x'^n} + \text{FF}(\text{LayerNorm}(\mathbf{x'^n}))
\end{equation}

Eqs.~\eqref{eq:2} and \eqref{eq:3} illustrate the Multi-Head Self-Attention (MHSA) layer and the feed-forward (FF) layer. The residual connection and pre-LayerNorm \cite{ba2016layer} are harnessed in both layers. The attention layer and feed-forward layer are formed as Transformer blocks and Transformer models are comprised of cascaded Transformer blocks. The class token is employed to classify the image and output the result.

ViT models leverage self-attention mechanisms to compute the similarity between each pair of patch tokens and then assign different weights to different tokens according to the similarities between the patch tokens. Nonetheless, ViT models often overlook the inductive bias inherent in images, where neighboring pixels or patches have more relations. This oversight can lead to slow convergence, requiring more training iterations to learn the inductive bias and demanding large datasets for optimal performance. In contrast, convolutions inherently possess an inductive bias due to local filters traversing the image, capturing local details. Recognizing the complementary nature of convolutions to Transformer models, particularly in scenarios with small datasets, we propose a lightweight approach using Depth-Wise Convolutions to enhance the convergence and performance of Vision Transformer models. This is particularly beneficial when training ViT models from scratch on limited datasets without additional assistance.

\begin{center}
\begin{figure*}[t]
    \centering
    \includegraphics[width=1.0\linewidth]{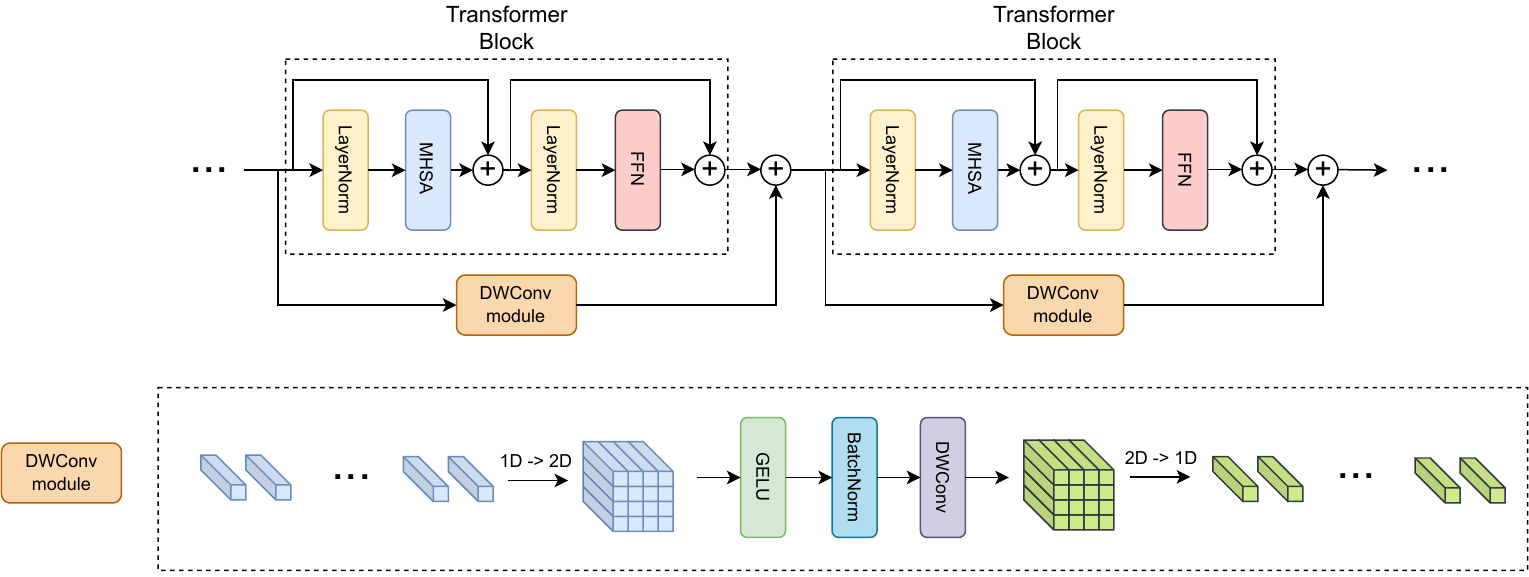}\hfill
    \caption{The architecture of our proposed method. The Depth-Wise Convolution module bypasses the entire Transformer block so that the local details can be attained and added to the output of the Transformer block. In the DWConv module, the 1D image patch tokens are first reshaped to 2D feature maps. If the class token exists, it would not be involved in the DWConv module and only image patch tokens are utilized to reconstruct the feature maps. Batch normalization and GELU activation are employed before the Depth-Wise Convolution. Finally, the feature maps would be reshaped to 1D tokens and added to the output of the Transformer block. The DWConv module is exploited in all Transformer blocks.}
	\label{fig:1}
\end{figure*}
\end{center}

\subsection{Our Approach}

Convolutional kernels excel at capturing fine details in images, a capability lacking in ViT models. The challenge lies in determining how and where to incorporate these kernels. To maintain a lightweight design without significantly increasing parameters and computational demands, we select Depth-Wise Convolutions to filter the local details. We utilize the Depth-Wise Convolution as the shortcut to bypass the entire Transformer block. Since the patch tokens are flattened to 1D, we have to reconstruct all patch tokens into 2D feature maps. The architecture of our proposed model is demonstrated in Fig.~\ref{fig:1}. The DWConv module is harnessed in all Transformer blocks as complementary components. 

$\mathbf{x^n}$ from Eq.~\eqref{eq:2} is used to reshape the 1D tokens to 2D feature maps. The reshaped 2D feature maps are implemented GELU activation \cite{hendrycks2016gaussian} and batch normalization \cite{ioffe2015batch} before being fed into the Depth-Wise Convolution (DWConv). The kernel we utilize for Depth-Wise Convolution is $3\times3$. The 2D feature maps are reshaped to 1D patch tokens and finally the reshaped 1D patch tokens ($\mathbf{{x_{1d}}^{n+1}}$) and the output of the Transformer block (Eq.~\eqref{eq:3}) are summed together. The summed result ($\mathbf{{x_{ours}}^{n+1}}$) is utilized as the input to the next block. This process is illustrated below.

\begin{equation}\label{eq:4}
    \mathbf{x_{2d}^n} = \text{Reshape}_{1d->2d}(\mathbf{x^n})
\end{equation}

\begin{equation}\label{eq:5}
    \mathbf{{x'_{2d}}^n} = \text{DWConv}(\text{BatchNorm}(\text{GELU}(\mathbf{{x_{2d}}^n})))
\end{equation}

\begin{equation}\label{eq:6}
    \mathbf{{x_{1d}}^{n+1}} = \text{Reshape}_{2d->1d}(\mathbf{{x'_{2d}}^n})
\end{equation}

\begin{equation}\label{eq:7}
    \mathbf{{x_{ours}}^{n+1}} = \mathbf{x^{n+1}} + \mathbf{{x_{1d}}^{n+1}}
\end{equation}

The DWConv modules act as ``supervisors" to supervise the Transformer blocks and they are complementary to each other. Each Transformer block is supervised by the DWConv modules to capture details that may be missed by the Transformer blocks. While the Transformer blocks play the main role in the architecture, the proposed lightweight DWConv modules are leveraged to retrieve local information, thereby enhancing the overall performance. Unlike some hybrid models that design complex hybrid architectures, our proposed approach demonstrates simplicity, effectiveness, and flexibility.

\begin{figure}[ht]
\begin{center}
   \includegraphics[width=0.8\linewidth]{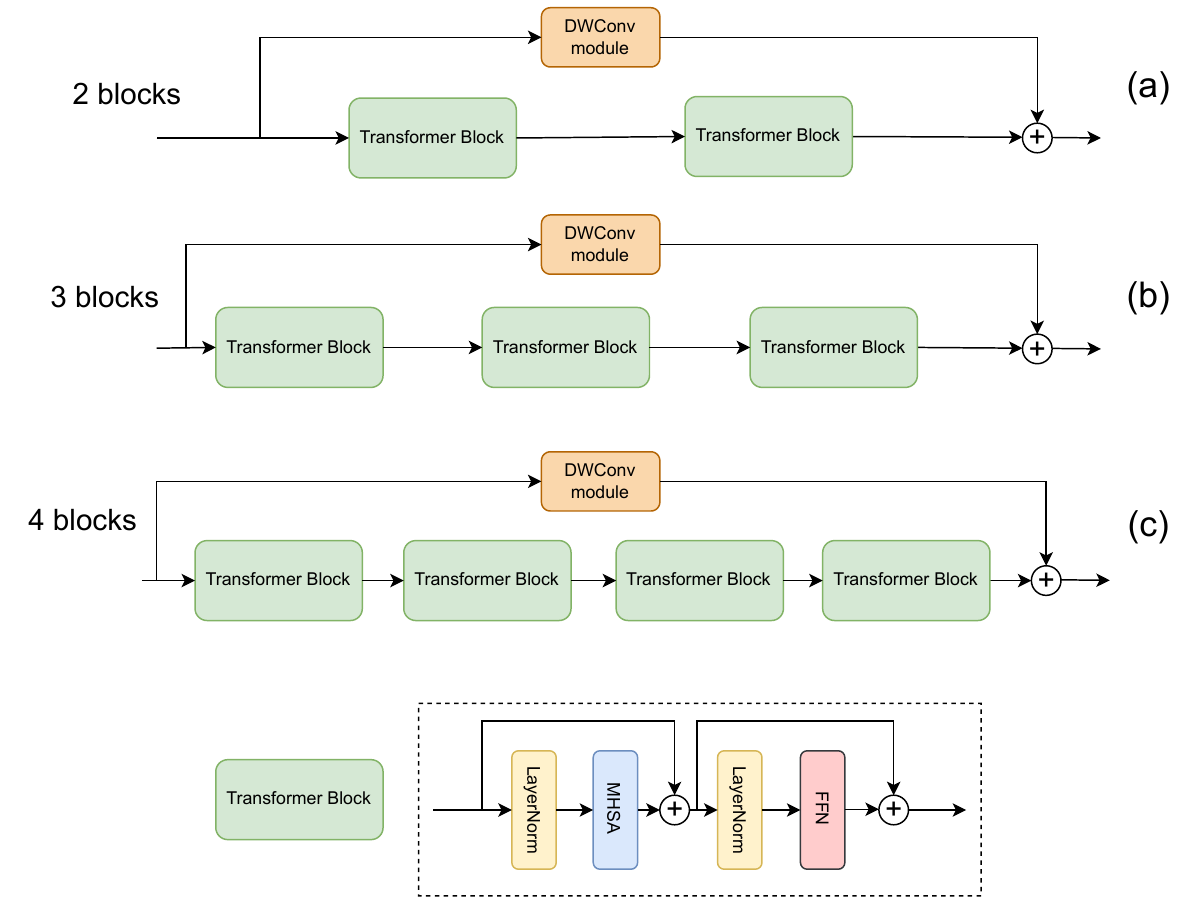}
\end{center}
   \caption{The architecture variants of our proposed approach involve bypassing multiple Transformer blocks. Structures (a), (b), and (c) represent the Depth-Wise module bypassing 2, 3, and 4 Transformer blocks, respectively. For Vision Transformer models with deeper layers, bypassing additional blocks may be a beneficial strategy to reduce both parameters and computational costs.}
\label{fig:2}
\end{figure}

\subsection{Architecture Variants}

In addition to the base architecture, we have designed several variants based on the core structure, as illustrated in Fig.~\ref{fig:2}. In our base architecture, the Depth-Wise module bypasses each Transformer block. Additional variants are designed where the Depth-Wise module bypasses more Transformer blocks. These variants prove beneficial when working with Vision Transformers that have deeper layers, helping to reduce the number of parameters and computational costs.

Moreover, in Transformer architectures with multiple stages, the size of the feature maps is reduced and the dimension is increased in successive stages. To maintain the input and output sizes of the Depth-Wise module, we recommend limiting the bypass within each stage to prevent a Depth-Wise module from crossing stages when multiple Transformer blocks are bypassed. Alternatively, one Depth-Wise module can be used to bypass an entire stage, ensuring that each stage has only one corresponding Depth-Wise module for more efficient combinations.

\begin{figure}[ht]
\begin{center}
   \includegraphics[width=0.8\linewidth]{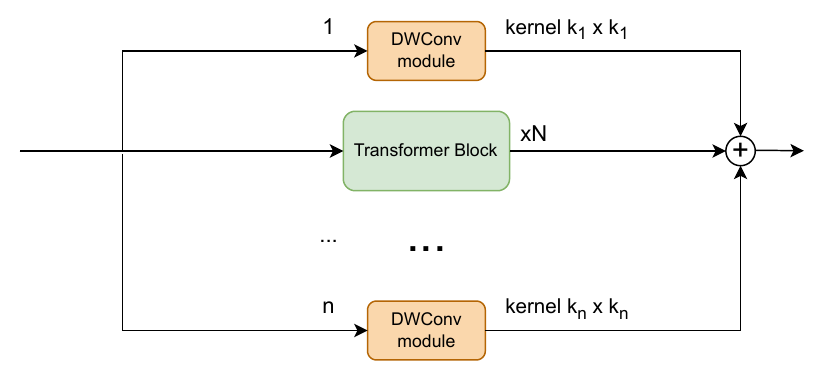}
\end{center}
   \caption{The architecture variants of our method involve multiple DWConv modules operating in parallel. These independent DWConv modules, each with different kernel sizes, run concurrently on Transformer blocks to capture local details simultaneously. This structure can be combined with previous variants shown in Fig.~\ref{fig:2} to include N Transformer blocks in the DWConv modules.}
\label{fig:3}
\end{figure}

Furthermore, the DWConv modules could operate in parallel with various kernel sizes to capture the local information independently, as illustrated in Fig.~\ref{fig:3}. In the experiments, we leverage parallel DWConv modules with different kernel sizes to demonstrate the performance of the variants. To further reduce the number of parameters and computational cost, multiple independent parallel DWConv modules could be combined with the aforementioned variants so that multiple Transformer blocks are contained by the DWConv modules. In Fig.~\ref{fig:3}, $N$ Transformer blocks are encompassed in the DWConv modules and $N \geq 1$.

Not modifying the structures of MHSA and FFN makes our approach more flexible for use with most Transformer models, rather than being designed for specific ones. The proposed architecture variants illustrate the flexibility of our methods compared to some existing hybrid architectures that combine convolutions and Transformers. The Transformers are greatly enhanced by the proposed DWConv module with minimal overhead. Additionally, the structure can be easily modified to further enhance the performance or save the parameters and computations with different variants.

\subsection{Complexity Analysis}

Our proposed approach is a lightweight module that is employed with each Transformer block. Unlike some models inserting convolutional layers inside the Transformer blocks, the proposed module is separable from the Transformer block, making it a plug-and-play module applicable to most existing Vision Transformer models. The increased parameters are dependent on the depths and dimensions of the Transformer models. Since our module is independent for each Transformer block without sharing parameters, deeper Transformer models could have more parameters introduced. However, the increased parameters are negligible compared to the Transformer backbone. 
For instance, the ViT-Tiny model used in our experiments has 12 blocks and a dimension of 192. With a depth-wise convolution of $3\times3$ kernel size, the increased parameters for the ViT-Tiny model are approximately $12\times192\times(3\times3 + 1) = 23, 040$ (0.023M) which is negligible compared to the backbone with around 5.5 million parameters. Moreover, since the patch size is 16 and the images are resized to 224 and the size of the feature maps is $14\times14$, the increased calculations for ViT-Tiny model could be approximately calculated by $12\times192\times(14\times14)\times(3\times3)=4,064,256$ (0.004G), which is trivial compared to the total 1.26G FLOPs. In the aforementioned calculation, the number of parameters and calculations of BatchNorm are ignored since they are insignificant to the model.

In the experiments, sometimes our methods could even reduce the number of parameters and FLOPs considering some modules and positional embeddings could be removed for training the small dataset when our approach is applied to the Vision Transformer models. The increased number of parameters and FLOPs that are trivial to the models are highly dependent on the number of layers and dimensions of the models. Additionally, they also depend on which architecture variants are employed for the Vision Transformer models. 

Some hybrid architectures merge convolutional networks into the Transformer architecture inefficiently, introducing significant parameters and computations as convolutional networks become essential components of Transformer structures. Additionally, these methods are often designed for specific Transformer architectures, making them impractical for other Transformer models. In contrast, our approach is designed to be easily incorporated into various Vision Transformer models. Complexity analysis shows that our approach introduces negligible overhead, with the majority of parameters and computations still coming from Transformer structures. However, the performance improvements are significant, especially on small datasets.

\section{Experiments and Results}

To verify the effectiveness and efficiency of our proposed approach, we select vanilla ViT \cite{dosovitskiy2020image}, CaiT \cite{touvron2021going}, and Swin Transformer \cite{liu2021swin} for the experiments on three small datasets: CIFAR10 \cite{krizhevsky2009learning}, CIFAR100 \cite{krizhevsky2009learning}, and Tiny-ImageNet \cite{le2015tiny}. We also evaluated the model on a relatively large dataset: ImageNet-1K \cite{russakovsky2015imagenet}. Additionally, COCO \cite{lin2014microsoft} is utilized for the evaluation of object detection and instance segmentation.

\subsection{Classification Performance on Small Datasets}

CIFAR10 \cite{krizhevsky2009learning}, CIFAR100 \cite{krizhevsky2009learning}, and Tiny-ImageNet \cite{le2015tiny} are exploited as small datasets for training and evaluating image classification tasks. The classification accuracy is defined as the ratio of correctly classified samples to the total number of samples. In our paper, we use Top-1 accuracy for classification. 

\subsubsection{Experimental Settings}

\textbf{ViT}.  We select ViT-Tiny and ViT-Small to conduct the experiments for all datasets. The parameter settings of ViT models are followed by \cite{steiner2021train}. The dimensions of ViT-Tiny and ViT-Small are 192 and 384, respectively. The MLP ratio is 4 for both models which indicates the MLP dimensions are 768 and 1536 for tiny and small models, respectively. The numbers of heads for tiny and small models in Multi-Head Self-Attention are 3 and 6 respectively so the dimension for each head is 64 for tiny and small models. The depths are 12 for ViT-Tiny and ViT-Small.

\textbf{CaiT}.  We choose CaiT-xxs12 and CaiT-xxs24 as the base models for the experiments. The dimensions of CaiT-xxs are 192 and the number of heads is 4. The MLP dimensions are 768 and the class depths are 2. The main depths for CaiT-xxs12 and CaiT-xxs24 are 12 and 24, respectively. 

\textbf{Swin Transformer}.  Swin-Tiny is selected for the experiments. For Swin-Tiny model, the drop path rate is 0.2 and the window size is 7; The depths and numbers of heads are (2, 2, 6, 2) and (3, 6, 12, 24) for each stage, respectively.

\textbf{Experimental Parameters}. All experiments are conducted using AdamW \cite{kingma2014adam} optimizer with 300 epochs and 20 epochs warmup. The weight decay is 0.05. The batch size for three small datasets is 128 with 4 NVIDIA P100 GPUs. The cosine decay learning rate scheduler is exploited. The base learning rate for Swin-Transformer on three small datasets is 2.5e-4, while the base learning rate for other experiments on small datasets is 5e-4. The images are resized to 224 and the patch size for both ViT and CaiT is 16.

\textbf{Data Augmentation}. Most regularization and augmentation settings follow \cite{liu2021swin}, including color jitter, Auto-Augment \cite{cubuk2018autoaugment}, random erasing \cite{zhong2020random}, MixUp \cite{zhang2017mixup}, CutMix \cite{yun2019cutmix}. All experiments are trained from scratch on each dataset without the assistance of an extra dataset.

\begin{table*}[t]
\centering
\setlength{\tabcolsep}{8pt}
\renewcommand{\arraystretch}{2.0}
\caption{The experimental results of ViT-Tiny and ViT-Small on small dataset (PE = Positional Embedding)}
\resizebox{\textwidth}{!}{
\begin{tabular}{ c|c|c|c|c|c|c|c|c|c}
 \hline
 \multirow{2}{*}{Model} & \multicolumn{3}{c|}{CIFAR-10} & 
     \multicolumn{3}{c|}{CIFAR-100} & \multicolumn{3}{c}{Tiny-ImageNet} \\
 \cline{2-10}
 & Accuracy & Params & FLOPs & Accuracy & Params & FLOPs & Accuracy & Params & FLOPs \\ 
 \hline

 \hline
ViT-Tiny & 94.01 & 5.5M & 1.26G & 73.68 & 5.5M & 1.26G & 59.00 & 5.6M & 1.26G \\

ViT-Tiny w/o PE & 87.83 (\textcolor{red}{-6.18}) & 5.5M & 1.26G & 64.41 (\textcolor{red}{-9.27}) & 5.5M & 1.26G & 53.15 (\textcolor{red}{-5.85}) & 5.5M & 1.26G \\

ViT-Tiny (\textbf{ours}) & \textbf{96.41} (\textcolor{blue}{+2.40}) & 5.5M & 1.26G & \textbf{78.05} (\textcolor{blue}{+4.37}) & 5.6M & 1.26G & \textbf{64.10} (\textcolor{blue}{+5.10}) & 5.6M & 1.26G \\

ViT-Tiny w/o PE (\textbf{ours}) & \textbf{96.32} (\textcolor{blue}{+2.31}) & 5.5M & 1.26G & \textbf{77.31} (\textcolor{blue}{+3.63}) & 5.5M & 1.26G & \textbf{63.57} (\textcolor{blue}{+4.57}) & 5.5M & 1.26G \\
 \hline
ViT-Small & 95.09 & 21.7M & 4.61G & 73.97 & 21.7M & 4.61G & 60.90 & 21.7M & 4.61G \\

ViT-Small w/o PE & 89.27 (\textcolor{red}{-5.82}) & 21.6M & 4.61G & 65.68 (\textcolor{red}{-8.29}) & 21.6M & 4.61G & 53.98 (\textcolor{red}{-6.92}) & 21.7M & 4.61G \\

ViT-Small (\textbf{ours}) & \textbf{97.02} (\textcolor{blue}{+1.93}) & 21.7M & 4.62G & \textbf{80.01} (\textcolor{blue}{+6.04}) & 21.7M & 4.62G & \textbf{66.86} (\textcolor{blue}{+5.96}) & 21.8M & 4.62G \\

ViT-Small w/o PE (\textbf{ours}) & \textbf{96.96} (\textcolor{blue}{+1.87}) & 21.6M & 4.62G & \textbf{80.14} (\textcolor{blue}{+6.17}) & 21.7M & 4.62G & \textbf{66.59} (\textcolor{blue}{+5.69}) & 21.7M & 4.62G \\
 \hline

 \end{tabular}
 }
 \label{table:vit}
 \end{table*}

\begin{table}[t]
\scriptsize
\centering
\setlength{\tabcolsep}{10pt}
\renewcommand{\arraystretch}{1.4}
\caption{The ablation study of ViT-Tiny (accuracy)}
{\begin{tabular}{c|c|c|c}
 \hline
Method & CIFAR-10 & CIFAR-100 & Tiny-ImageNet\\
 \hline
shortcut & 87.50 & 65.10 & 52.15 \\
kernel 3 & 96.32 & 77.31 & 63.57 \\
kernel 5 & 96.26 & \textbf{78.71} & 63.67 \\
kernel 7 & 96.26 & 78.69 & 63.95 \\
kernel 3+5 & \textbf{96.52} & 78.63 & 64.00 \\
kernel 3+5+7 & 96.39 & 78.00 & \textbf{64.27} \\
\hline

\end{tabular}
}
\label{table:vit_alation}
\end{table}

\subsubsection{Vision Transformer}

The vanilla ViT model splits the image into small patches which are embedded as tokens for Transformer blocks. Since the tokens are 1-dimensional without the 2-dimensional positional information, the vanilla ViT model utilizes a positional embedding which would be added to all tokens to learn the 2-dimensional positional relationship between tokens. Since convolutions with zero padding could encode the positional information\cite{chu2021conditional}, we also applied our approach to the ViT model without positional embeddings. The experimental results are illustrated in Table~\ref{table:vit}.

From Table~\ref{table:vit} we observe that removing the positional embeddings significantly deteriorates the performance of ViT-Tiny and ViT-Small, highlighting the importance of positional embeddings in Vision Transformers. When our method is applied to ViT models, there is a substantial improvement in performance, regardless of the presence of positional embeddings. For ViT-Tiny, the increased accuracy for CIFAR-10, CIFAR-100, and Tiny-ImageNet are around 2\%, 4\%, and 5\%, respectively. For ViT-Small, the performance boost for CIFAR-10, CIFAR-100, and Tiny-ImageNet are nearly 2\%, 6\%, and 6\%, respectively. Additionally, our method without positional embeddings even slightly reduces the number of parameters with much better accuracy. More importantly, the accuracy of ViT-Tiny with our proposed DWConv surpasses that of vanilla ViT-Small which has almost 4x the number of parameters and FLOPs by large margins, demonstrating the efficiency and effectiveness of our proposed method.

Moreover, extra experiments are implemented with different kernel sizes and parallel DWConv modules, as illustrated in Table~\ref{table:vit_alation}. Directly applying a shortcut connection with positional embedding without any modules to bypass the Transformer blocks significantly reduces the accuracy. The possible reason for that might be the low-level input embeddings for the Transformers, which is different from the high-level input features attained from CNNs \cite{ma2021miti}.
Some large kernel sizes or parallel DWConv modules could boost the performance with a slightly higher number of parameters and FLOPs.

\subsubsection{CaiT}

CaiT introduces LayerScale to improve the performance of deeper layer transformer models by multiplying a learnable diagonal matrix \cite{touvron2021going} by each residual block. The class attention is introduced before the final classifier to convert patch embeddings into the final class embeddings. Talking heads attention \cite{shazeer2020talking} is utilized in the model for further improvement of the performance. However, LayerScale \cite{touvron2021going} and talking heads attention \cite{shazeer2020talking} are not necessary when our proposed approach is applied to CaiT model on a small dataset. Moreover, talking heads attention is extremely time-consuming for small dataset training in our experiments. Thus LayerScale and talking heads attention are removed when our method is applied to CaiT-xxs12 and CaiT-xxs24, which is demonstrated in Table~\ref{table:cait}, where ``LS", ``TH" and ``PE" illustrate LayerScale, talking heads attention and positional embeddings. Similar to the vanilla ViT model, the positional embeddings are not necessary when our method is introduced to the small dataset training.

\begin{table*}[t]
\centering
\setlength{\tabcolsep}{8pt}
\renewcommand{\arraystretch}{2.0}
\caption{The experimental results of CaiT-xxs12 and CaiT-xxs24 on small dataset (LS = LayerScale, TH = Talking Head, PE = Positional Embedding)}
\resizebox{\textwidth}{!}{
\begin{tabular}{ c|c|c|c|c|c|c|c|c|c}
 \hline
 \multirow{2}{*}{Model} & \multicolumn{3}{c|}{CIFAR-10} & 
     \multicolumn{3}{c|}{CIFAR-100} & \multicolumn{3}{c}{Tiny-ImageNet} \\
 \cline{2-10}
 & Accuracy & Params & FLOPs & Accuracy & Params & FLOPs & Accuracy & Params & FLOPs \\ 
 \hline

 \hline
CaiT-xxs12 & 92.02 & 6.4M & 1.30G & 73.43 & 6.4M & 1.30G & 59.17 & 6.5M & 1.30G \\

CaiT-xxs12 w/o-(LS, TH, PE) & 87.45 (\textcolor{red}{-4.57}) & 6.4M & 1.28G & 70.32 (\textcolor{red}{-3.11}) & 6.4M & 1.28G & 60.62 (\textcolor{blue}{+1.45}) & 6.4M & 1.28G \\

CaiT-xxs12  w/o-(LS, TH, PE) (\textbf{ours}) & \textbf{96.43} (\textcolor{blue}{+4.41}) & 6.4M & 1.29G & \textbf{81.72} (\textcolor{blue}{+8.29}) & 6.4M & 1.29G & \textbf{70.47} (\textcolor{blue}{+11.30}) & 6.4M & 1.29G \\
\hline

CaiT-xxs24 & 93.89 & 11.8M & 2.53G & 74.84 & 11.8M & 2.53G & 60.97 & 11.8M & 2.53G \\

CaiT-xxs24 w/o-(LS, TH, PE) (\textbf{ours}) & \textbf{97.28} (\textcolor{blue}{+3.39}) & 11.8M & 2.51G & \textbf{82.83} (\textcolor{blue}{+7.99}) & 11.8M & 2.51G & \textbf{70.64} (\textcolor{blue}{+9.67}) & 11.8M & 2.51G \\
 \hline

 \end{tabular}
 }
 \label{table:cait}
 \end{table*}

\begin{table}[t]
\scriptsize
\centering
\setlength{\tabcolsep}{10pt}
\renewcommand{\arraystretch}{1.4}
\caption{The accuracy for different blocks bypassed by DWConv module with CaiT}
{\begin{tabular}{c|c|c|c}
 \hline
Method & CIFAR-10 & CIFAR-100 & Tiny-ImageNet\\
\hline
xxs12 (1 block) & \textbf{96.43} & \textbf{81.72} & \textbf{70.47} \\
xxs12 (2 blocks) & 96.16 & 80.67 & 68.76 \\
xxs12 (3 blocks) & 95.60 & 79.50 & 67.61 \\
xxs12 (4 blocks) & 94.80 & 78.61 & 67.49 \\

\hline
xxs24 (1 block) & \textbf{97.28} & 82.83 & \textbf{70.64} \\
xxs24 (2 blocks) & 97.00 & \textbf{82.90} & 70.07 \\
xxs24 (3 blocks) & 96.84 & 81.90 & 69.57 \\
xxs24 (4 blocks) & 96.51 & 80.16 & 68.61 \\

\hline

\end{tabular}
}
\label{table:cait_alation}
\end{table}

It is evident from Table~\ref{table:cait} that removing LayerScale, talking heads attention, and positional embeddings reduces the accuracy for small datasets except Tiny-ImageNet. When our DWConv modules are applied to CaiT models, the accuracy is significantly boosted for small datasets. For Tiny-ImageNet, the accuracy of CaiT-xxs12 with our proposed DWConv is tremendously improved by around 11\% with less number of parameters and FLOPs since the aforementioned modules are eliminated when our approach is applied to CaiT models. Similar to ViT models, CaiT-xxs12 with our method has much higher accuracy than the original CaiT-xxs24 and almost half of the number of parameters and FLOPs compared to CaiT-xxs24 model. In addition, CaiT-xxs12 with our DWConv modules even has much better performance than the original Swin-Transformer (as shown in Table~\ref{table:swin}) that has almost 4x the number of parameters and FLOPs than CaiT-xxs12.

To verify the architecture variant that multiple Transformer blocks are bypassed by the proposed DWConv modules, more experiments are conducted with CaiT, as demonstrated in Table~\ref{table:cait_alation}. The number of blocks indicates how many Transformer blocks are supervised by the DWConv modules in the architecture. The performance drops when more blocks are supervised by DWConv modules, but the accuracy is still much higher than the original models. The variant is appropriate when the layers are deeper to reduce the number of parameters and FLOPs while still maintaining relatively high accuracy.

\begin{table*}[t]
\centering
\setlength{\tabcolsep}{8pt}
\renewcommand{\arraystretch}{2.0}
\caption{The experimental results of Swin-Tiny on small datasets}
\resizebox{\textwidth}{!}{
\begin{tabular}{ c|c|c|c|c|c|c|c|c|c}
 \hline
 \multirow{2}{*}{Model} & \multicolumn{3}{c|}{CIFAR-10} & 
     \multicolumn{3}{c|}{CIFAR-100} & \multicolumn{3}{c}{Tiny-ImageNet} \\
 \cline{2-10}
 & Accuracy & Params & FLOPs & Accuracy & Params & FLOPs & Accuracy & Params & FLOPs \\ 
 \hline

Swin-Tiny & 93.59 & 27.5M & 4.51G & 78.75 & 27.6M & 4.51G & 68.24 & 27.7M & 4.51G \\

Swin-Tiny w/o shift-window & 93.36 (\textcolor{red}{-0.23}) & 27.5M & 4.51G & 78.51 (\textcolor{red}{-0.24}) & 27.6M & 4.51G & 68.30 (\textcolor{blue}{+0.06}) & 27.7M & 4.51G \\

Swin-Tiny (\textbf{ours}) & \textbf{96.92} (\textcolor{blue}{+3.33}) & 27.6M & 4.52G & \textbf{83.92} (\textcolor{blue}{+5.17}) & 27.6M & 4.52G & \textbf{71.96} (\textcolor{blue}{+3.72}) & 27.7M & 4.52G \\

Swin-Tiny  w/o shift-window (\textbf{ours}) & \textbf{97.18} (\textcolor{blue}{+3.59}) & 27.6M & 4.52G & \textbf{83.38} (\textcolor{blue}{+4.63}) & 27.6M & 4.52G & \textbf{72.36} (\textcolor{blue}{+4.12}) & 27.7M & 4.52G \\

Swin-Tiny kernel 3+5 (\textbf{ours}) & \textbf{97.06} (\textcolor{blue}{+3.47}) & 27.7M & 4.56G & \textbf{83.84} (\textcolor{blue}{+5.09}) & 27.8M & 4.56G & \textbf{72.74} (\textcolor{blue}{+4.50}) & 27.8M & 4.56G \\

\hline

 \end{tabular}
 }
 \label{table:swin}
 \end{table*}

\begin{center}
\begin{figure}[t]
    \centering
    \includegraphics[width=1\linewidth]{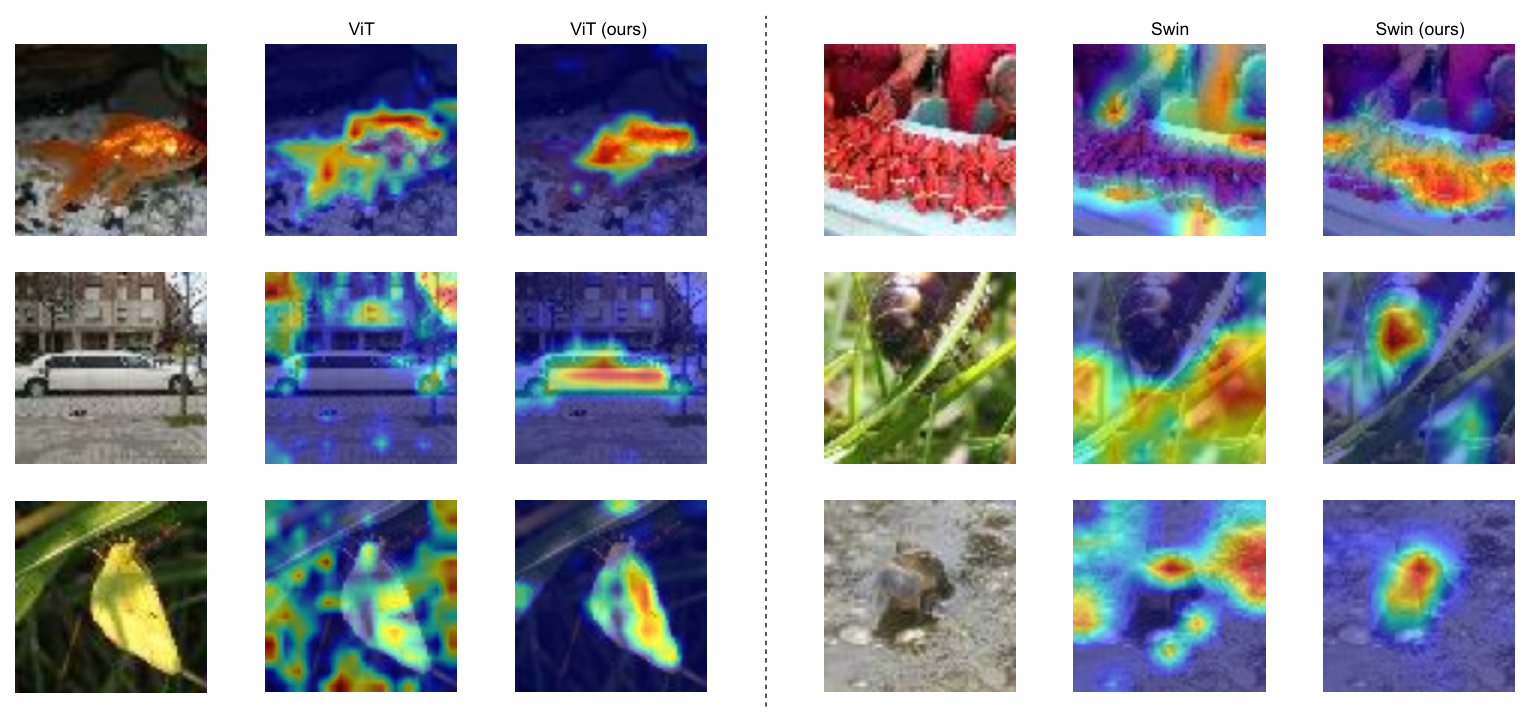}\hfill
    \caption{Some Grad-CAM visualization with ViT-Tiny and Swin-Tiny models. The vanilla Transformer models tend to capture the global information, as illustrated in the CAM visualizations. With our method, the models are able to capture both local details and global perspectives, particularly when dealing with smaller objects. Please note that the original images in the figure are from the Tiny-ImageNet dataset with a low resolution of 64$\times$64 pixels. Thus, they appear blurred when enlarged.}
\label{fig:4}
\end{figure}
\end{center}

\subsubsection{Swin Transformer}

The architecture of Swin Transformer consists of four stages with hierarchical feature maps. The size of feature maps is reduced by 2 on each side by merging adjacent image patches in the following successive stage. Shifted window-based self-attention \cite{liu2021swin} is proposed to extend the view of the tokens instead of limiting the view of the tokens in the windows they are assigned. In the experiments, we investigate the effectiveness of our method applied to Swin Transformer, which is demonstrated in Table~\ref{table:swin}.

\begin{figure}[t!]
\centering
\includegraphics[width=.45\textwidth]{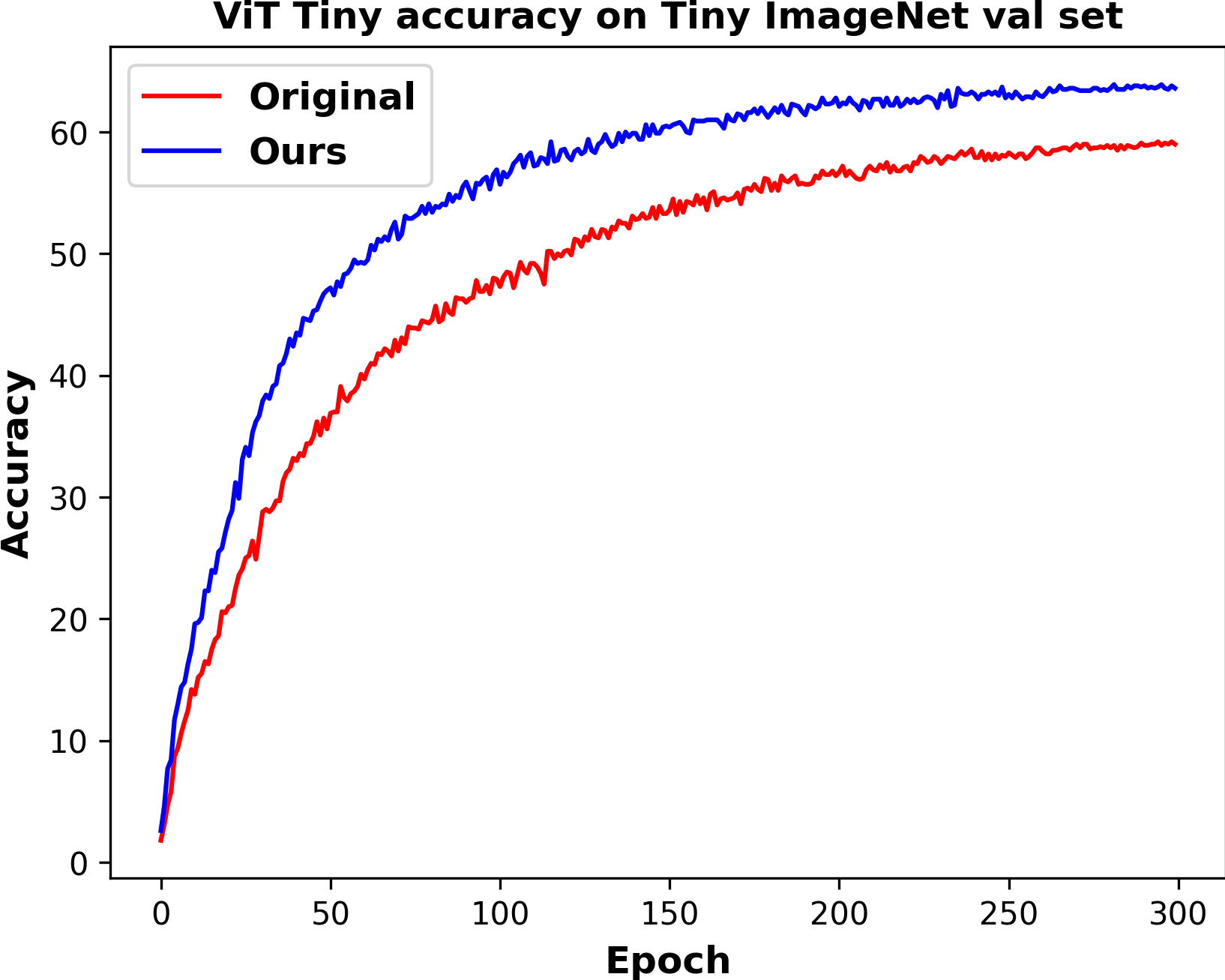}
\includegraphics[width=.45\textwidth]{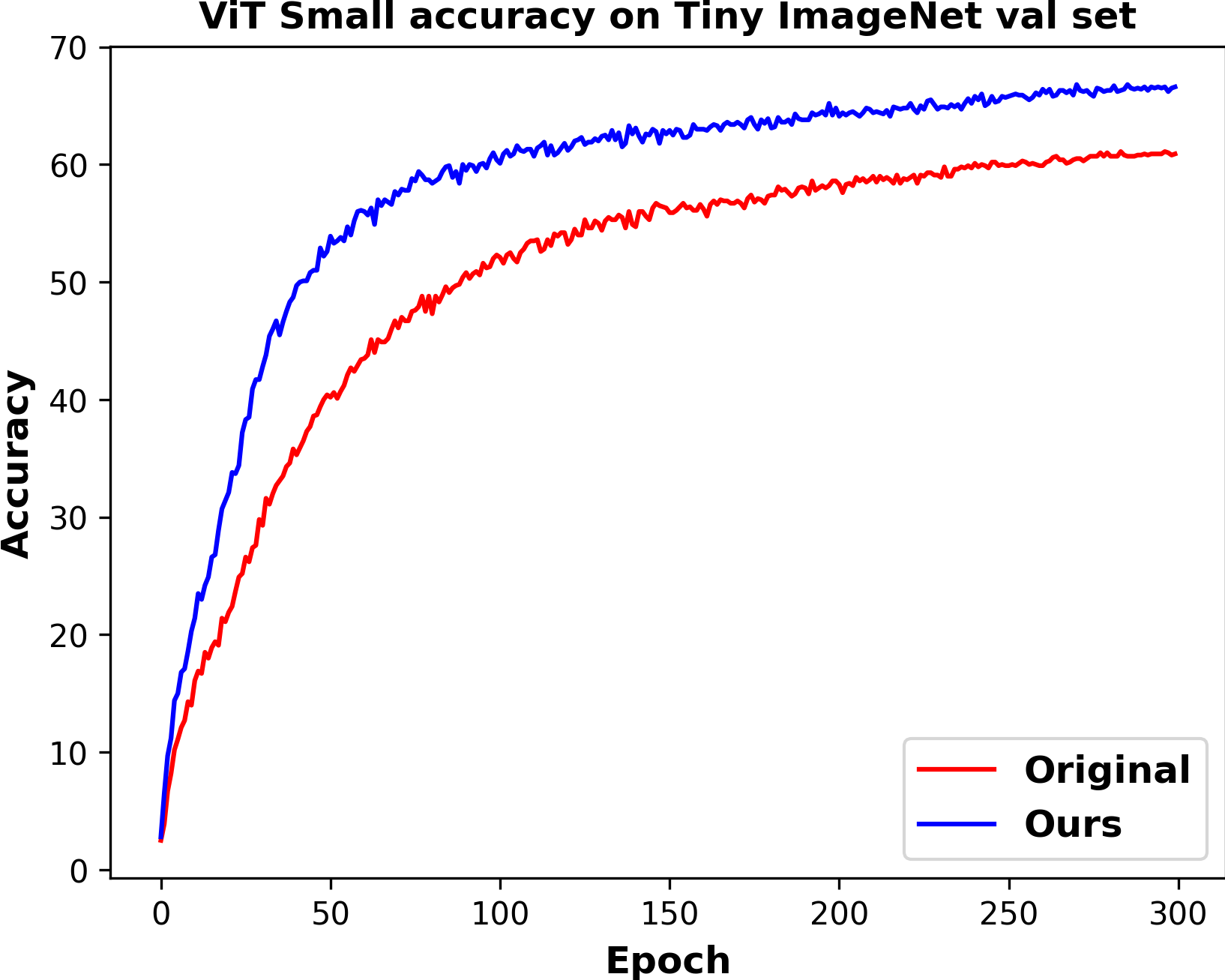} \\
\includegraphics[width=.45\textwidth]{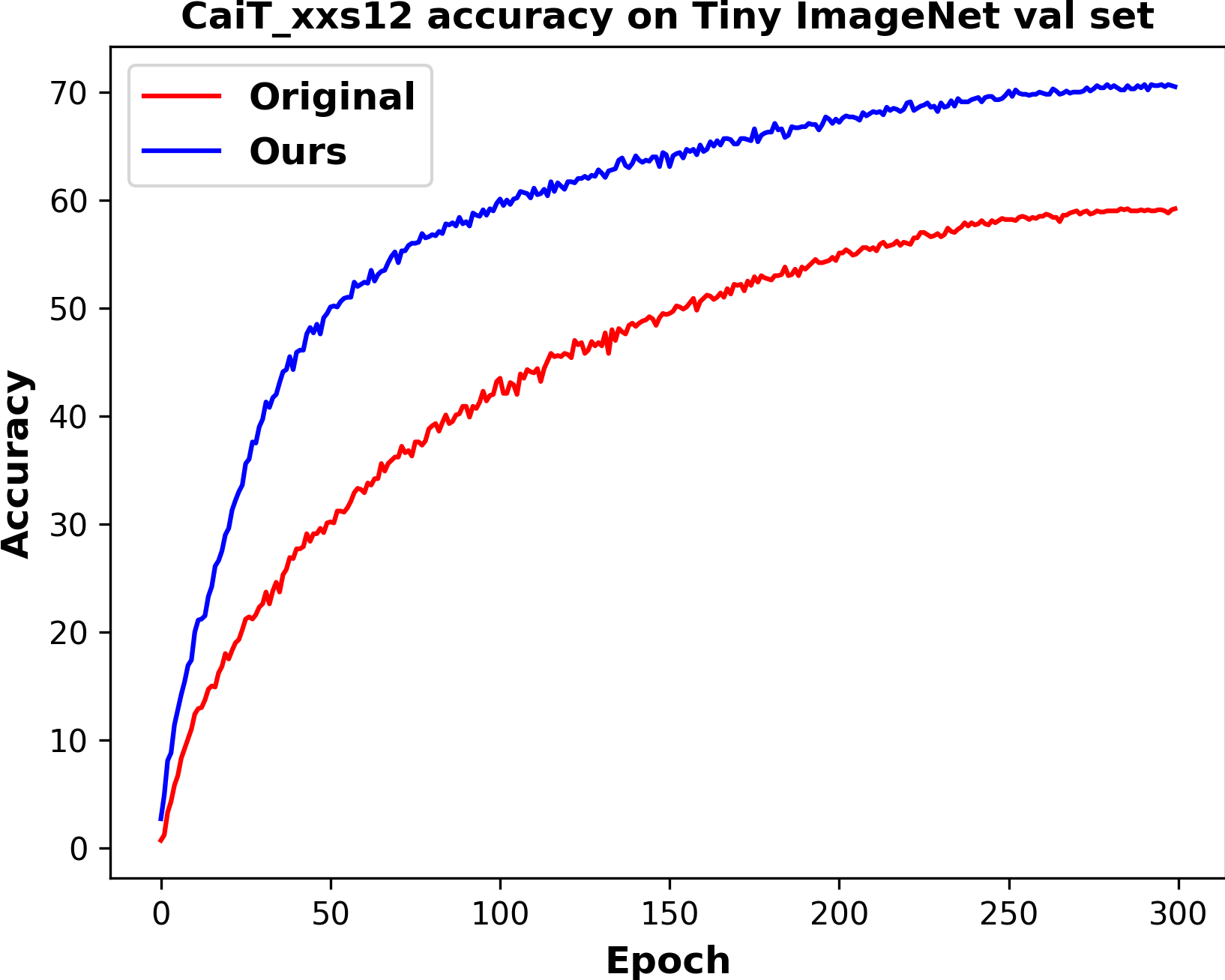}
\includegraphics[width=.45\textwidth]{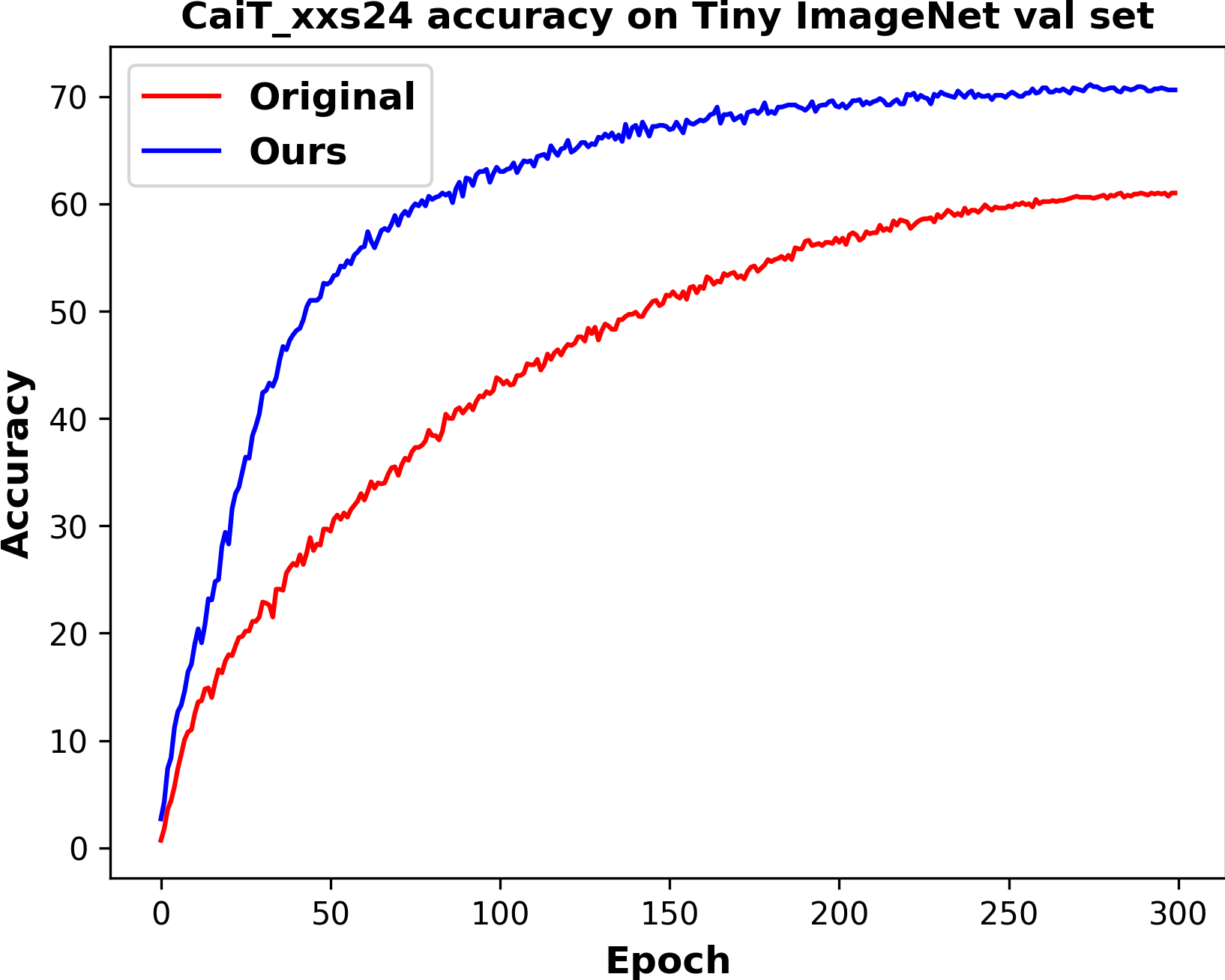} \\
\includegraphics[width=.45\textwidth]{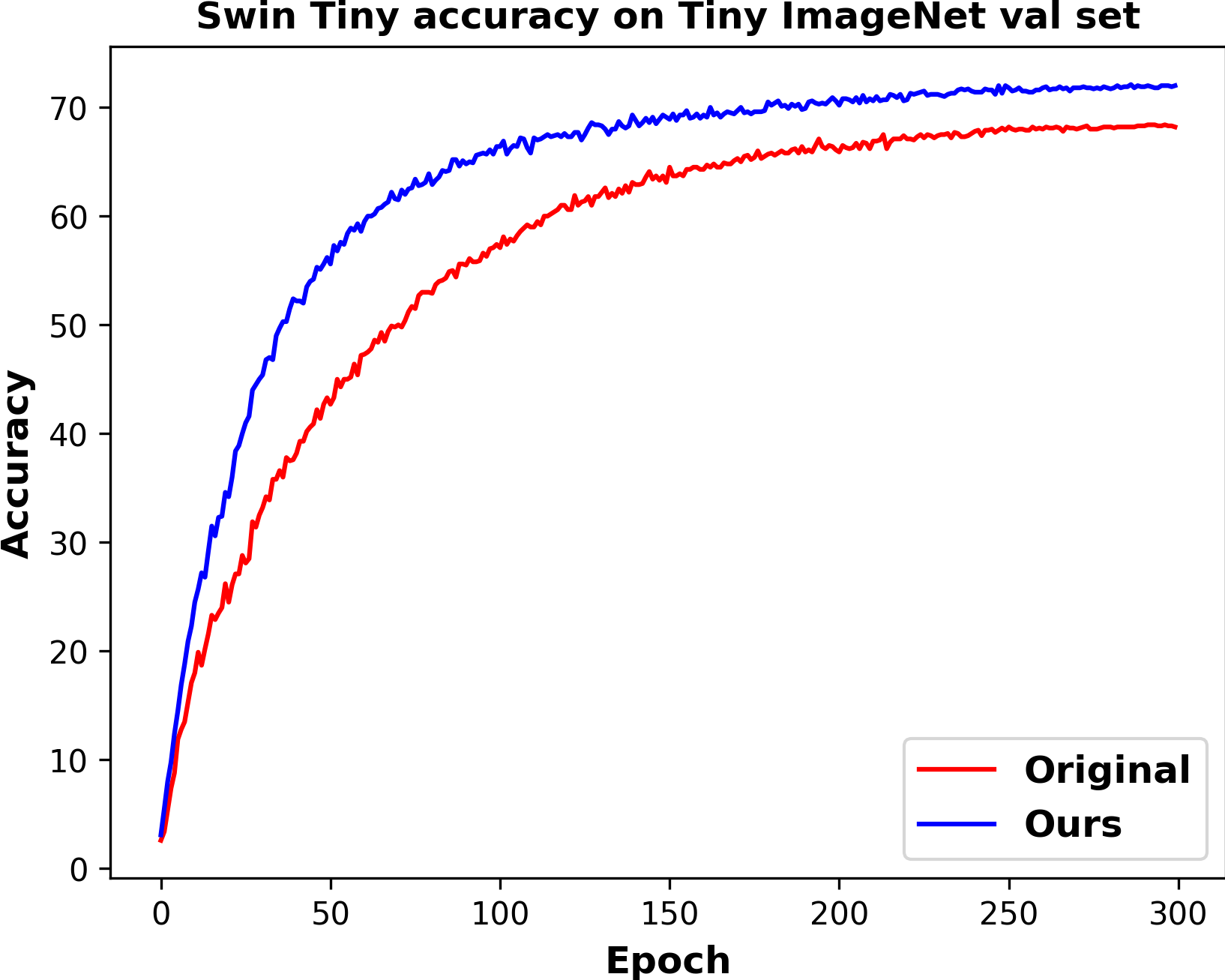}
\caption{The accuracy for val set during the training on Tiny-ImageNet for 300 epochs. The blue curves indicate our method and the red curves are from the original models. The accuracy for val set is recorded for each epoch. The convergence of the models with our approach is much faster than the original models.}
\label{fig:5}
\end{figure}

The shifted window approach does not have too much effect on the performance of small datasets. Swin Transformer model with our method has much better accuracy than the original model with negligible parameter overhead. 
In addition, ``kernel 3+5" means parallel DWConv modules have kernel size 3 and 5, respectively. Independent parallel DWConv modules with different kernels increase the accuracy in some cases, but the number of parameters and computations would be slightly increased.

We also utilize GRAD-CAM \cite{selvaraju2017grad} to visualize the focus areas of the models, as depicted in Fig.~\ref{fig:4}. The Transformer models exhibit global views of images, while Transformer models might overlook some objects due to a lack of local information, especially when the objects are relatively small. With our method, both global and local information could be captured and enhanced by each other, which could improve the performance of the models.

The convergence of our approach is significantly faster than the original models, which is demonstrated in Fig.~\ref{fig:5}. The accuracy on val set is recorded for each epoch on Tiny-ImageNet for all models. Our method exhibits much higher performance and considerably faster convergence speed. Our approach could reach a similar accuracy at around or less than 100 epochs while the original models require 300 epochs to attain the same accuracy. Similar performance curves are observed for CIFAR-10 and CIFAR-100.

\subsection{Classification Performance on ImageNet-1K}

In addition to the small datasets, we have also evaluated the models on a relatively large dataset, ImageNet-1K \cite{russakovsky2015imagenet}, to further verify the effectiveness of our approach. ImageNet-1K \cite{russakovsky2015imagenet} contains nearly 1.3 million images for training and 50k images for validation. For ImageNet-1K \cite{russakovsky2015imagenet}, The batch size is 1024 with 8 NVIDIA V100 GPUs and the base learning rate is 1e-3.

\begin{table}[t]
\scriptsize
\centering
\setlength{\tabcolsep}{15pt}
\renewcommand{\arraystretch}{1.7}
\caption{The performance on ImageNet-1K}
{\begin{tabular}{c|c|c c|c}
 \hline
Model & Accuracy & Params & FLOPs & Year\\
 \hline
ResNet-18 \cite{he2016deep}\cite{wightman2021resnet} & 71.5 & 11.7M & 1.8G & 2016 \\
ResNet-34 \cite{he2016deep}\cite{wightman2021resnet} & 76.4 & 21.8M & 3.7G & 2016 \\
ResNet-50 \cite{he2016deep}\cite{wightman2021resnet} & 80.4 & 25.6M & 4.1G & 2016 \\
SE-ResNet-50 \cite{hu2018squeeze}\cite{wightman2021resnet} & 80.0 & 28.1M & 4.1G & 2018 \\ 
 \hline
DeiT-Ti \cite{touvron2021training}\cite{zhang2023vitaev2} & 72.2 & 5.7M & 1.3G & 2021 \\
Visformer-Ti \cite{chen2021visformer} & 78.6 & 10.3M & 1.3G & 2021 \\
PVT-Tiny \cite{wang2021pyramid} & 75.1 & 13.2M & 1.9G & 2021 \\
DeiT-S \cite{touvron2021training}\cite{zhang2023vitaev2} & 79.9 & 22.1M & 4.6G & 2021 \\
PVT-Small \cite{wang2021pyramid} & 79.8 & 24.5M & 3.8G & 2021 \\
MSG-T \cite{fang2022msg} & 80.9 & 28M & 4.6G & 2022 \\
DiT-B1 \cite{ma2023dit} & 79.9 & 30.3M & 2.0G & 2023 \\
Visformer-S \cite{chen2021visformer} & 81.5 & 40.2M & 4.9G & 2021 \\
 \hline
CaiT-xxs12 & 74.85 & 6.6M & 1.3G & - \\
CaiT-xxs24 & 77.66 & 11.9M & 2.5G & - \\
Swin-Tiny & 81.14 & 28.3M & 4.5G & - \\
\hline
CaiT-xxs12 (\textbf{ours}) & \textbf{75.89} (\textcolor{blue}{+1.04}) & 6.6M & 1.3G & - \\
CaiT-xxs24 (\textbf{ours}) & \textbf{79.66} (\textcolor{blue}{+2.00}) & 12.0M & 2.5G & - \\
Swin-Tiny (\textbf{ours}) & \textbf{81.73} (\textcolor{blue}{+0.59}) & 28.3M & 4.5G & - \\
\hline

\end{tabular}
}
\label{table:imagenet}
\end{table}

We utilize CaiT \cite{touvron2021going} and Swin Transformer \cite{liu2021swin} to illustrate the performance of our approach on ImageNet. The kernel size for our method is $3\times3$ and the DWConv module is applied to each Transformer block. When our approach is applied to the models, the positional embeddings and talking heads attention in CaiT are retained for better accuracy, while LayerScale is eliminated. For Swin Transformer, our approach is directly applied to the model without any other changes. The experimental results are demonstrated in Table~\ref{table:imagenet}. We employ top-1 accuracy to measure the performance of the models and the results of ResNet models are extracted from \cite{wightman2021resnet}. The performance of CaiT and Swin-Transformer on ImageNet-1K is further boosted (up to 2\%) by our method.

Moreover, in comparison to the convolutional counterparts like ResNet \cite{he2016deep}, our approach still has superior performance with insignificant parameters and FLOPs overhead. Especially when the layers of the Transformer models go deeper (e.g., CaiT-xxs24), the improvement is even higher on ImageNet.

\begin{figure}[t]
\begin{center}
   \includegraphics[width=0.8\linewidth]{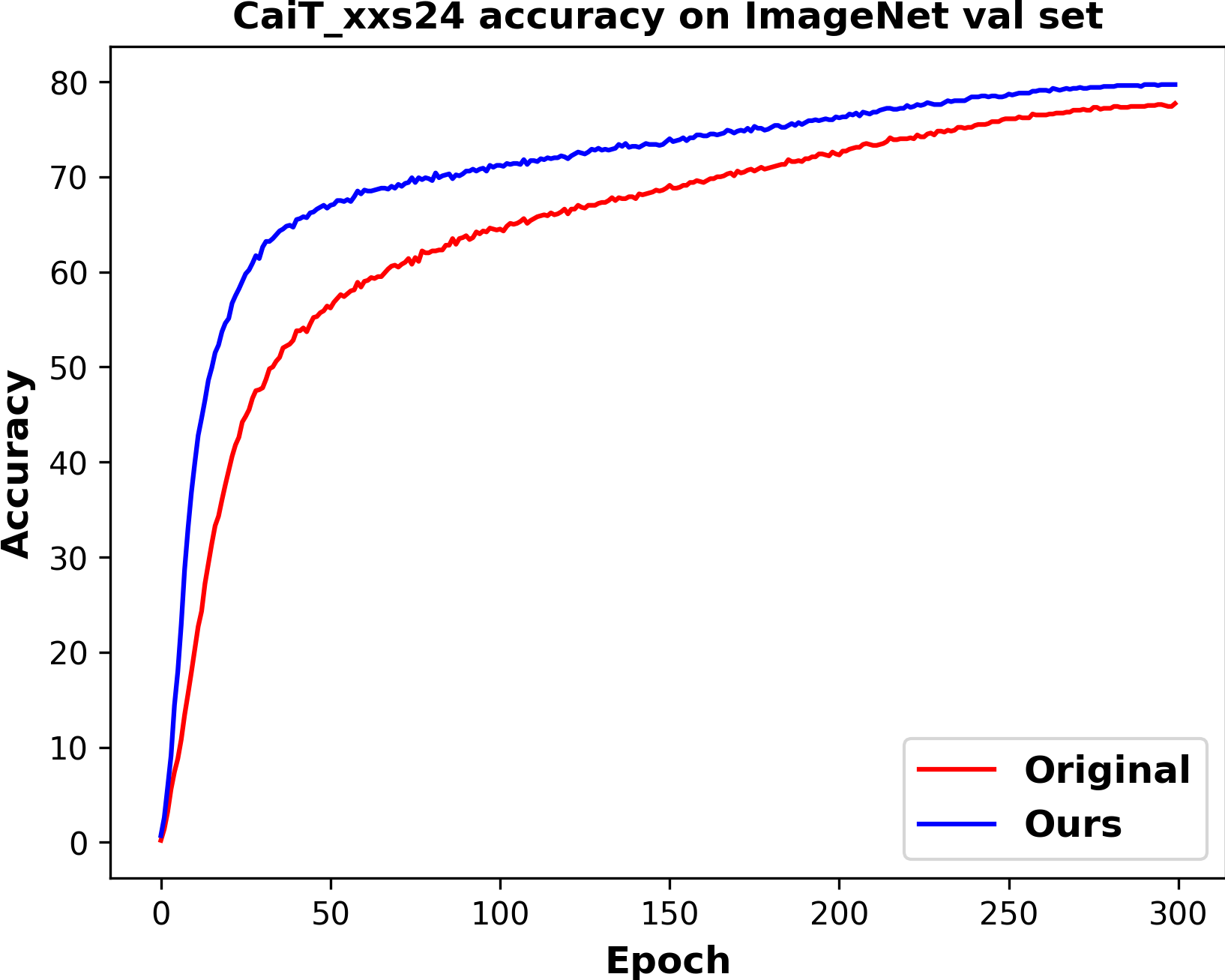}
\end{center}
   \caption{The accuracy of CaiT-xxs24 for val set during the training on ImageNet for 300 epochs. The blue curve demonstrates our method. The convergence rate for our approach is much faster than the original model.}
\label{fig:imagenet}
\end{figure}

We also visualize the convergence curve of CaiT-xxs24 on ImageNet. As illustrated in Fig.~\ref{fig:imagenet}, the convergence rate of our approach is much faster than the original model by a large margin when the epoch is less than 100. This experiment indicates that our proposed approach could achieve higher accuracy with significantly faster convergence speed on a relatively large dataset.

\begin{table}[t]
\scriptsize
\centering
\setlength{\tabcolsep}{10pt}
\renewcommand{\arraystretch}{1.6}
\caption{Experiments for Object Detection and Instance Segmentation}
{\begin{tabular}{c||c|c c||c|c c}
 \hline
  \multirow{2}{*}{Model} & \multicolumn{3}{c||}{Object Detection} & 
     \multicolumn{3}{c}{Instance Segmentation} \\
 \cline{2-7}
 & $mAP$ & $AP_{50}$ & $AP_{75}$ & $mAP$ & $AP_{50}$ & $AP_{75}$ \\
 \hline
Mask-RCNN  & 28.2 & 48.3 & 29.3 & 27.4 & 45.6 & 28.7 \\
Mask-RCNN (\textbf{ours})  & \textbf{30.6} & \textbf{50.2} & \textbf{32.6} & \textbf{28.9} & \textbf{47.4} & \textbf{30.6}  \\
\hline
Cas Mask-RCNN  & 35.3 & 52.5 & 38.0 & 31.4 & 49.8 & 33.7 \\
Cas Mask-RCNN (\textbf{ours})  & \textbf{36.2} & \textbf{53.0} & \textbf{39.1} & \textbf{32.1} & \textbf{50.6} & \textbf{34.5} \\
\hline

\end{tabular}
}
\label{table:det_seg}
\end{table}

\subsection{Object Detection and Instance Segmentation}

In addition to Image Classification, we apply the proposed approach to object detection and instance segmentation and conduct the experiments on COCO dataset \cite{lin2014microsoft} with Mask RCNN \cite{he2017mask} and Cascade Mask R-CNN \cite{cai2018cascade}. The backbone utilized in the experiments is Swin-Tiny \cite{liu2021swin}. The models are trained from scratch without pre-trained backbones.

For the experimental settings, we employ AdamW \cite{kingma2014adam} as the optimizer and the warmup iterations are 500. We utilize four NVIDIA P100 GPUs to train the model with 2 samples for each GPU. The initial learning rate is set at 5e-5 and the learning rate is reduced by 10 at epochs 9 and 12, respectively. The image scale for the experiments is 1333$\times$800. The total epochs for the experiments are 12.

\begin{figure}[t]
\centering
\includegraphics[width=.13\textwidth]{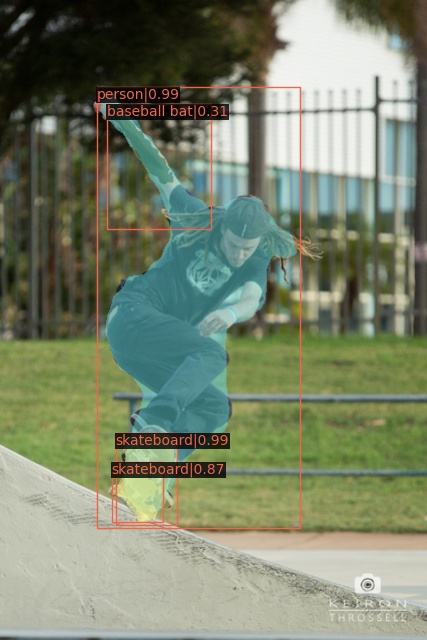}
\includegraphics[width=.292\textwidth]{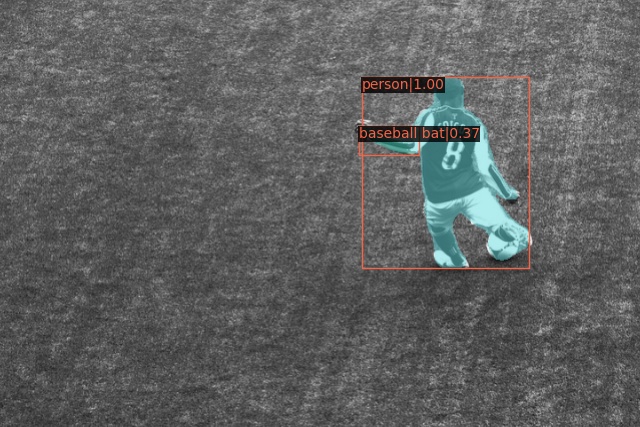}
\includegraphics[width=.26\textwidth]{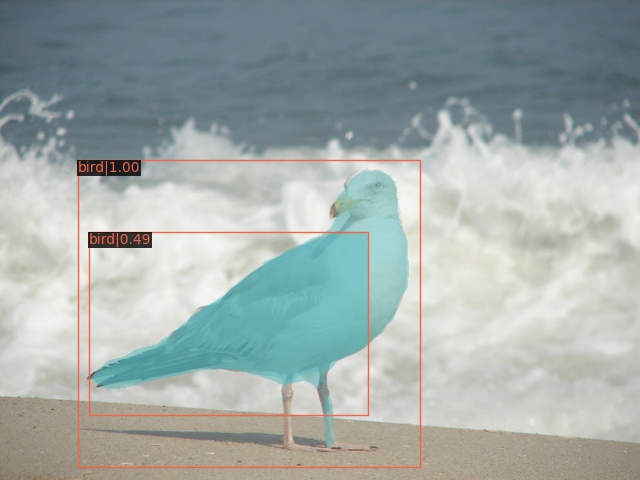}
\includegraphics[width=.26\textwidth]{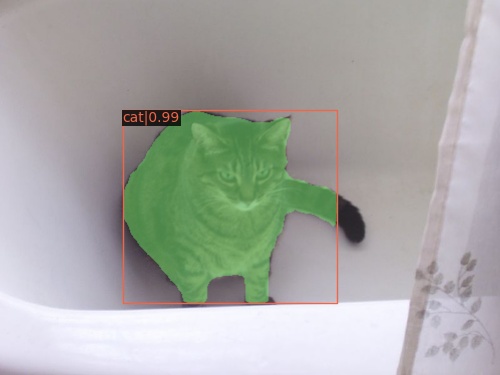} \\
\includegraphics[width=.13\textwidth]{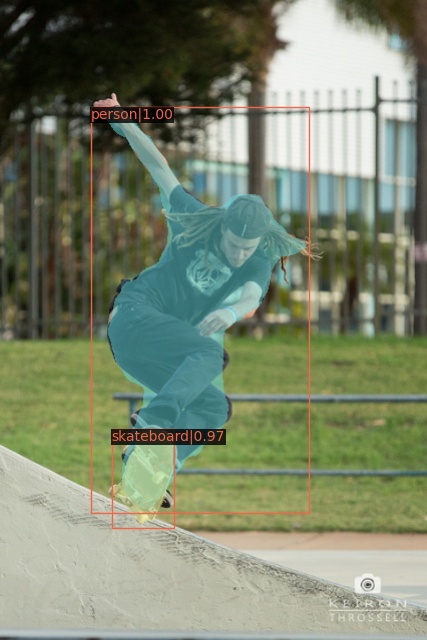}
\includegraphics[width=.292\textwidth]{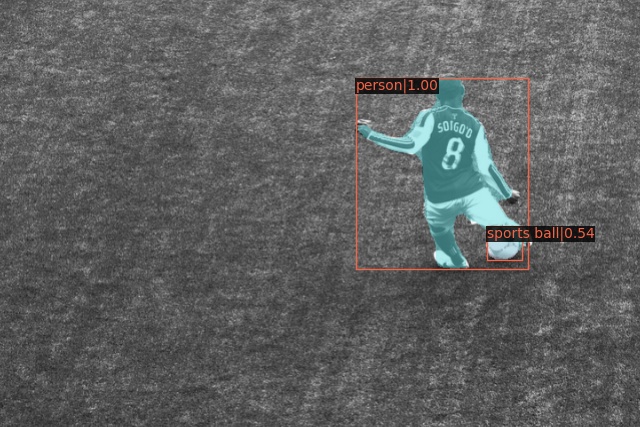}
\includegraphics[width=.26\textwidth]{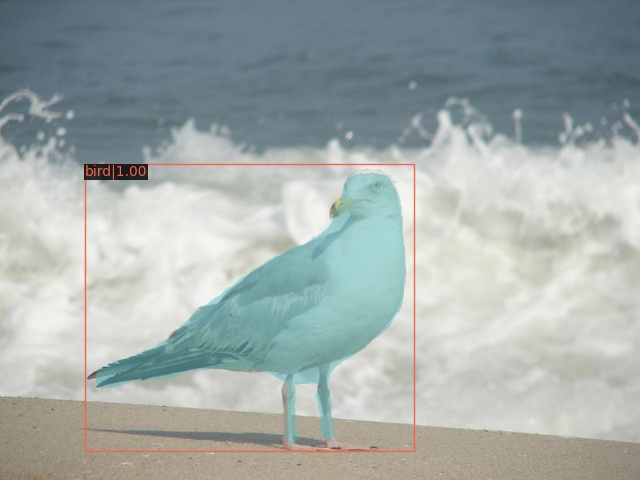}
\includegraphics[width=.26\textwidth]{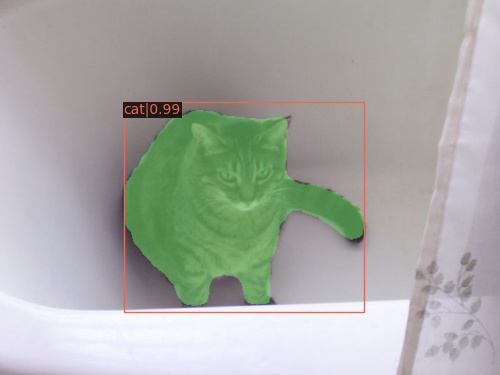}
\caption{The visualization of object detection and instance segmentation between the original method (the first row) and ours (the second row) with Mask-RCNN. 
The results demonstrate that our approach better detects small objects and produces more accurately predicted boundaries for the objects.}
\label{fig:7}
\end{figure}

The results for the experiments of object detection and instance segmentation are illustrated in Table~\ref{table:det_seg}, where ``Cas Mask-RCNN" stands for ``Cascade Mask-RCNN". From the definition in COCO \cite{lin2014microsoft}, ``mAP" refers to the average precision results that are averaged over all classes. The average precision is calculated by averaging the results over IoU thresholds from 0.5 to 0.95 with a step of 0.05. Additionally, ``$AP_{50}$" represents the average precision computed using only the IoU threshold of 0.5 and ``$AP_{75}$" indicates the average precision computed using only the IoU threshold of 0.75. Additionally, the visualization of the original method and our proposed approach with Mask-RCNN \cite{he2017mask} is illustrated in Fig.~\ref{fig:7}. In Fig.~\ref{fig:7}, the first row demonstrates the visualization results of the original model and the second row indicates the visualization results of our proposed model.

The experimental results clearly show that our approach improves the performance of backbone networks for object detection and instance segmentation, demonstrating the effectiveness of our proposed method across various vision tasks. The effectiveness likely stems from the fine-detailed information captured by the proposed DWConv module. Object detection and instance segmentation require detailed information for predicting object boundaries and pixel-level labels, respectively. Vision Transformers might lack the ability to capture extensive fine-detailed information, especially when used as the backbone. Our proposed DWConv module complements this limitation with minimal overhead.

\subsection{Analysis}

The proposed DWConv module, which bypasses the entire Transformer block, demonstrates higher accuracy for image classification, object detection, and instance segmentation when the models are trained from scratch. Additionally, this module enables Transformer models to achieve a much faster convergence rate for image classification, especially on relatively small datasets. Furthermore, our approach can significantly enhance small-size Transformer models, even surpassing large-size original Transformer models with substantially more parameters and computations on small datasets for image classification. Our approach illustrates both the effectiveness and efficiency of Vision Transformer models. 

Although our architecture performs well on relatively small datasets due to the inductive bias introduced by the DWConv module, the improvement might not be as pronounced when the dataset is relatively large. The abundant data can mitigate the drawbacks of Transformer models.
In addition, since the proposed DWConv module is lightweight and plug-and-play, it may not show significant improvement when models have a large number of parameters and computations. A large number of parameters and computations can increase the representation ability of Transformer models and potentially remedy the lack of inductive bias, albeit inefficiently. However, our proposed models enhance both the effectiveness and efficiency of Transformer models, achieving higher accuracy than some large models, despite having significantly fewer parameters and computations.

Moreover, our method may not show significant improvement for transfer learning. One strength of our proposed approach is that our light-weight module can be utilized in most Transformer models, potentially enhancing performance, particularly on small datasets, when training from scratch. The experiments in this paper are all conducted with training from scratch. The possible reason for the lack of significant improvement in transfer learning is that pre-trained models already possess substantial representation ability, reducing the necessity for the inductive bias introduced by our approach. Thus, our proposed module may not provide much additional benefit when pre-trained models are applied to other tasks or datasets for fine-tuning.

\section{Conclusion}

In this paper, we have presented a straightforward yet impactful approach that utilizes Depth-Wise Convolution modules to bypass Transformer blocks, enabling Vision Transformer models to capture both global and local information with minimal overhead. Extensive experimental evaluations show that small Transformer models, when equipped with our method, outperform larger Transformer models with significantly more parameters and FLOPs on small datasets for image classification. Our approach also significantly improves performance on ImageNet-1K \cite{russakovsky2015imagenet} for classification and COCO \cite{lin2014microsoft} for object detection and instance segmentation when trained from scratch. Additionally, we introduce several architecture variants tailored to different models and objectives. We anticipate that our method will inspire further research on Vision Transformers, particularly in the context of small datasets.

\section*{Acknowledgements}
    The project is partly supported by the Natural Sciences and Engineering Research Council of Canada (NSERC) under grant numbers 1-51-48183 and 1-51-48933.



  \bibliographystyle{elsarticle-num} 
  \bibliography{reference}

\begin{thebibliography}{10}
\expandafter\ifx\csname url\endcsname\relax
  \def\url#1{\texttt{#1}}\fi
\expandafter\ifx\csname urlprefix\endcsname\relax\def\urlprefix{URL }\fi
\expandafter\ifx\csname href\endcsname\relax
  \def\href#1#2{#2} \def\path#1{#1}\fi

\bibitem{vaswani2017attention}
A.~Vaswani, N.~Shazeer, N.~Parmar, J.~Uszkoreit, L.~Jones, A.~N. Gomez, {\L}.~Kaiser, I.~Polosukhin, Attention is all you need, Advances in neural information processing systems 30 (2017).

\bibitem{dosovitskiy2020image}
A.~Dosovitskiy, L.~Beyer, A.~Kolesnikov, D.~Weissenborn, X.~Zhai, T.~Unterthiner, M.~Dehghani, M.~Minderer, G.~Heigold, S.~Gelly, et~al., An image is worth 16x16 words: Transformers for image recognition at scale, arXiv preprint arXiv:2010.11929 (2020).

\bibitem{howard2017mobilenets}
A.~G. Howard, M.~Zhu, B.~Chen, D.~Kalenichenko, W.~Wang, T.~Weyand, M.~Andreetto, H.~Adam, Mobilenets: Efficient convolutional neural networks for mobile vision applications, arXiv preprint arXiv:1704.04861 (2017).

\bibitem{patel2022aggregating}
K.~Patel, A.~M. Bur, F.~Li, G.~Wang, Aggregating global features into local vision transformer, in: 2022 26th International Conference on Pattern Recognition (ICPR), IEEE, 2022, pp. 1141--1147.

\bibitem{chen2023accumulated}
X.~Chen, Q.~Hu, K.~Li, C.~Zhong, G.~Wang, Accumulated trivial attention matters in vision transformers on small datasets, in: Proceedings of the IEEE/CVF Winter Conference on Applications of Computer Vision, 2023, pp. 3984--3992.

\bibitem{zhu2023biformer}
L.~Zhu, X.~Wang, Z.~Ke, W.~Zhang, R.~W. Lau, Biformer: Vision transformer with bi-level routing attention, in: Proceedings of the IEEE/CVF conference on computer vision and pattern recognition, 2023, pp. 10323--10333.

\bibitem{wang2023pst}
Y.~Wang, H.~Zhu, G.~Wang, Pst-net: Point cloud completion network based on local geometric feature reuse and neighboring recovery with taylor approximation, in: 2023 International Joint Conference on Neural Networks (IJCNN), IEEE, 2023, pp. 1--8.

\bibitem{sajid2021audio}
U.~Sajid, X.~Chen, H.~Sajid, T.~Kim, G.~Wang, Audio-visual transformer based crowd counting, in: Proceedings of the IEEE/CVF International Conference on Computer Vision, 2021, pp. 2249--2259.

\bibitem{touvron2021training}
H.~Touvron, M.~Cord, M.~Douze, F.~Massa, A.~Sablayrolles, H.~J{\'e}gou, Training data-efficient image transformers \& distillation through attention, in: International conference on machine learning, PMLR, 2021, pp. 10347--10357.

\bibitem{touvron2021going}
H.~Touvron, M.~Cord, A.~Sablayrolles, G.~Synnaeve, H.~J{\'e}gou, Going deeper with image transformers, in: Proceedings of the IEEE/CVF international conference on computer vision, 2021, pp. 32--42.

\bibitem{wang2021pyramid}
W.~Wang, E.~Xie, X.~Li, D.-P. Fan, K.~Song, D.~Liang, T.~Lu, P.~Luo, L.~Shao, Pyramid vision transformer: A versatile backbone for dense prediction without convolutions, in: Proceedings of the IEEE/CVF international conference on computer vision, 2021, pp. 568--578.

\bibitem{heo2021rethinking}
B.~Heo, S.~Yun, D.~Han, S.~Chun, J.~Choe, S.~J. Oh, Rethinking spatial dimensions of vision transformers, in: Proceedings of the IEEE/CVF International Conference on Computer Vision, 2021, pp. 11936--11945.

\bibitem{chen2021psvit}
B.~Chen, P.~Li, B.~Li, C.~Li, L.~Bai, C.~Lin, M.~Sun, J.~Yan, W.~Ouyang, Psvit: Better vision transformer via token pooling and attention sharing, arXiv preprint arXiv:2108.03428 (2021).

\bibitem{chen2022improving}
X.~Chen, Y.~Qin, W.~Xu, A.~M. Bur, C.~Zhong, G.~Wang, Improving vision transformers on small datasets by increasing input information density in frequency domain, in: IEEE/CVF International Conference on Computer Vision Workshops (ICCVW), Vol.~2, 2022.

\bibitem{chen2024superlora}
X.~Chen, J.~Liu, Y.~Wang, M.~Brand, G.~Wang, T.~Koike-Akino, et~al., Superlora: Parameter-efficient unified adaptation of multi-layer attention modules, arXiv preprint arXiv:2403.11887 (2024).

\bibitem{liu2021swin}
Z.~Liu, Y.~Lin, Y.~Cao, H.~Hu, Y.~Wei, Z.~Zhang, S.~Lin, B.~Guo, Swin transformer: Hierarchical vision transformer using shifted windows, in: Proceedings of the IEEE/CVF international conference on computer vision, 2021, pp. 10012--10022.

\bibitem{fang2022msg}
J.~Fang, L.~Xie, X.~Wang, X.~Zhang, W.~Liu, Q.~Tian, Msg-transformer: Exchanging local spatial information by manipulating messenger tokens, in: Proceedings of the IEEE/CVF Conference on Computer Vision and Pattern Recognition, 2022, pp. 12063--12072.

\bibitem{chu2021twins}
X.~Chu, Z.~Tian, Y.~Wang, B.~Zhang, H.~Ren, X.~Wei, H.~Xia, C.~Shen, Twins: Revisiting the design of spatial attention in vision transformers, Advances in Neural Information Processing Systems 34 (2021) 9355--9366.

\bibitem{huang2021shuffle}
Z.~Huang, Y.~Ben, G.~Luo, P.~Cheng, G.~Yu, B.~Fu, Shuffle transformer: Rethinking spatial shuffle for vision transformer, arXiv preprint arXiv:2106.03650 (2021).

\bibitem{wu2021cvt}
H.~Wu, B.~Xiao, N.~Codella, M.~Liu, X.~Dai, L.~Yuan, L.~Zhang, Cvt: Introducing convolutions to vision transformers, in: Proceedings of the IEEE/CVF international conference on computer vision, 2021, pp. 22--31.

\bibitem{srinivas2021bottleneck}
A.~Srinivas, T.-Y. Lin, N.~Parmar, J.~Shlens, P.~Abbeel, A.~Vaswani, Bottleneck transformers for visual recognition, in: Proceedings of the IEEE/CVF conference on computer vision and pattern recognition, 2021, pp. 16519--16529.

\bibitem{li2021localvit}
Y.~Li, K.~Zhang, J.~Cao, R.~Timofte, L.~Van~Gool, Localvit: Bringing locality to vision transformers, arXiv preprint arXiv:2104.05707 (2021).

\bibitem{guo2022cmt}
J.~Guo, K.~Han, H.~Wu, Y.~Tang, X.~Chen, Y.~Wang, C.~Xu, Cmt: Convolutional neural networks meet vision transformers, in: Proceedings of the IEEE/CVF Conference on Computer Vision and Pattern Recognition, 2022, pp. 12175--12185.

\bibitem{chen2022mobile}
Y.~Chen, X.~Dai, D.~Chen, M.~Liu, X.~Dong, L.~Yuan, Z.~Liu, Mobile-former: Bridging mobilenet and transformer, in: Proceedings of the IEEE/CVF Conference on Computer Vision and Pattern Recognition, 2022, pp. 5270--5279.

\bibitem{lu2022bridging}
Z.~Lu, H.~Xie, C.~Liu, Y.~Zhang, Bridging the gap between vision transformers and convolutional neural networks on small datasets, Advances in Neural Information Processing Systems 35 (2022) 14663--14677.

\bibitem{xu2021vitae}
Y.~Xu, Q.~Zhang, J.~Zhang, D.~Tao, Vitae: Vision transformer advanced by exploring intrinsic inductive bias, Advances in neural information processing systems 34 (2021) 28522--28535.

\bibitem{zhang2023vitaev2}
Q.~Zhang, Y.~Xu, J.~Zhang, D.~Tao, Vitaev2: Vision transformer advanced by exploring inductive bias for image recognition and beyond, International Journal of Computer Vision (2023) 1--22.

\bibitem{chen2022mixformer}
Q.~Chen, Q.~Wu, J.~Wang, Q.~Hu, T.~Hu, E.~Ding, J.~Cheng, J.~Wang, Mixformer: Mixing features across windows and dimensions, in: Proceedings of the IEEE/CVF conference on computer vision and pattern recognition, 2022, pp. 5249--5259.

\bibitem{wei2023dmformer}
Z.~Wei, H.~Pan, L.~Li, M.~Lu, X.~Niu, P.~Dong, D.~Li, Dmformer: Closing the gap between cnn and vision transformers, in: ICASSP 2023-2023 IEEE International Conference on Acoustics, Speech and Signal Processing (ICASSP), IEEE, 2023, pp. 1--5.

\bibitem{nie2024scopevit}
X.~Nie, H.~Jin, Y.~Yan, X.~Chen, Z.~Zhu, D.~Qi, Scopevit: Scale-aware vision transformer, Pattern Recognition 153 (2024) 110470.

\bibitem{su2024dctvit}
K.~Su, L.~Cao, B.~Zhao, N.~Li, D.~Wu, X.~Han, Y.~Liu, Dctvit: Discrete cosine transform meet vision transformers, Neural Networks 172 (2024) 106139.

\bibitem{jamali2023wetmapformer}
A.~Jamali, S.~K. Roy, P.~Ghamisi, Wetmapformer: A unified deep cnn and vision transformer for complex wetland mapping, International Journal of Applied Earth Observation and Geoinformation 120 (2023) 103333.

\bibitem{wang2022hybrid}
Q.~Wang, Y.~Liu, Z.~Xiong, Y.~Yuan, Hybrid feature aligned network for salient object detection in optical remote sensing imagery, IEEE transactions on geoscience and remote sensing 60 (2022) 1--15.

\bibitem{liu2023transcending}
Y.~Liu, Z.~Xiong, Y.~Yuan, Q.~Wang, Transcending pixels: boosting saliency detection via scene understanding from aerial imagery, IEEE Transactions on Geoscience and Remote Sensing (2023).

\bibitem{liu2023local}
F.~Liu, Y.~Kong, L.~Zhang, G.~Feng, B.~Yin, Local-global coordination with transformers for referring image segmentation, Neurocomputing 522 (2023) 39--52.

\bibitem{ba2016layer}
J.~L. Ba, J.~R. Kiros, G.~E. Hinton, Layer normalization, arXiv preprint arXiv:1607.06450 (2016).

\bibitem{hendrycks2016gaussian}
D.~Hendrycks, K.~Gimpel, Gaussian error linear units (gelus), arXiv preprint arXiv:1606.08415 (2016).

\bibitem{ioffe2015batch}
S.~Ioffe, C.~Szegedy, Batch normalization: Accelerating deep network training by reducing internal covariate shift, in: International conference on machine learning, pmlr, 2015, pp. 448--456.

\bibitem{krizhevsky2009learning}
A.~Krizhevsky, G.~Hinton, et~al., Learning multiple layers of features from tiny images (2009).

\bibitem{le2015tiny}
Y.~Le, X.~Yang, Tiny imagenet visual recognition challenge, CS 231N 7~(7) (2015) 3.

\bibitem{russakovsky2015imagenet}
O.~Russakovsky, J.~Deng, H.~Su, J.~Krause, S.~Satheesh, S.~Ma, Z.~Huang, A.~Karpathy, A.~Khosla, M.~Bernstein, et~al., Imagenet large scale visual recognition challenge, International journal of computer vision 115 (2015) 211--252.

\bibitem{lin2014microsoft}
T.-Y. Lin, M.~Maire, S.~Belongie, J.~Hays, P.~Perona, D.~Ramanan, P.~Doll{\'a}r, C.~L. Zitnick, Microsoft coco: Common objects in context, in: Computer Vision--ECCV 2014: 13th European Conference, Zurich, Switzerland, September 6-12, 2014, Proceedings, Part V 13, Springer, 2014, pp. 740--755.

\bibitem{steiner2021train}
A.~Steiner, A.~Kolesnikov, X.~Zhai, R.~Wightman, J.~Uszkoreit, L.~Beyer, How to train your vit? data, augmentation, and regularization in vision transformers, arXiv preprint arXiv:2106.10270 (2021).

\bibitem{kingma2014adam}
D.~P. Kingma, J.~Ba, Adam: A method for stochastic optimization, arXiv preprint arXiv:1412.6980 (2014).

\bibitem{cubuk2018autoaugment}
E.~D. Cubuk, B.~Zoph, D.~Mane, V.~Vasudevan, Q.~V. Le, Autoaugment: Learning augmentation policies from data, arXiv preprint arXiv:1805.09501 (2018).

\bibitem{zhong2020random}
Z.~Zhong, L.~Zheng, G.~Kang, S.~Li, Y.~Yang, Random erasing data augmentation, in: Proceedings of the AAAI conference on artificial intelligence, Vol.~34, 2020, pp. 13001--13008.

\bibitem{zhang2017mixup}
H.~Zhang, M.~Cisse, Y.~N. Dauphin, D.~Lopez-Paz, mixup: Beyond empirical risk minimization, arXiv preprint arXiv:1710.09412 (2017).

\bibitem{yun2019cutmix}
S.~Yun, D.~Han, S.~J. Oh, S.~Chun, J.~Choe, Y.~Yoo, Cutmix: Regularization strategy to train strong classifiers with localizable features, in: Proceedings of the IEEE/CVF international conference on computer vision, 2019, pp. 6023--6032.

\bibitem{chu2021conditional}
X.~Chu, Z.~Tian, B.~Zhang, X.~Wang, X.~Wei, H.~Xia, C.~Shen, Conditional positional encodings for vision transformers, arXiv preprint arXiv:2102.10882 (2021).

\bibitem{ma2021miti}
W.~Ma, T.~Zhang, G.~Wang, Miti-detr: Object detection based on transformers with mitigatory self-attention convergence, arXiv preprint arXiv:2112.13310 (2021).

\bibitem{shazeer2020talking}
N.~Shazeer, Z.~Lan, Y.~Cheng, N.~Ding, L.~Hou, Talking-heads attention, arXiv preprint arXiv:2003.02436 (2020).

\bibitem{selvaraju2017grad}
R.~R. Selvaraju, M.~Cogswell, A.~Das, R.~Vedantam, D.~Parikh, D.~Batra, Grad-cam: Visual explanations from deep networks via gradient-based localization, in: Proceedings of the IEEE international conference on computer vision, 2017, pp. 618--626.

\bibitem{he2016deep}
K.~He, X.~Zhang, S.~Ren, J.~Sun, Deep residual learning for image recognition, in: Proceedings of the IEEE conference on computer vision and pattern recognition, 2016, pp. 770--778.

\bibitem{wightman2021resnet}
R.~Wightman, H.~Touvron, H.~J{\'e}gou, Resnet strikes back: An improved training procedure in timm, arXiv preprint arXiv:2110.00476 (2021).

\bibitem{hu2018squeeze}
J.~Hu, L.~Shen, G.~Sun, Squeeze-and-excitation networks, in: Proceedings of the IEEE conference on computer vision and pattern recognition, 2018, pp. 7132--7141.

\bibitem{chen2021visformer}
Z.~Chen, L.~Xie, J.~Niu, X.~Liu, L.~Wei, Q.~Tian, Visformer: The vision-friendly transformer, in: Proceedings of the IEEE/CVF international conference on computer vision, 2021, pp. 589--598.

\bibitem{ma2023dit}
Y.~Ma, Z.~Fei, J.~Huang, Dit: Efficient vision transformers with dynamic token routing, arXiv preprint arXiv:2308.03409 (2023).

\bibitem{he2017mask}
K.~He, G.~Gkioxari, P.~Doll{\'a}r, R.~Girshick, Mask r-cnn, in: Proceedings of the IEEE international conference on computer vision, 2017, pp. 2961--2969.

\bibitem{cai2018cascade}
Z.~Cai, N.~Vasconcelos, Cascade r-cnn: Delving into high quality object detection, in: Proceedings of the IEEE conference on computer vision and pattern recognition, 2018, pp. 6154--6162.

\end{thebibliography}





\end{document}